\documentclass{article}

\usepackage{lineno}
\usepackage{pdflscape}
\usepackage{array}
\usepackage[longtable]{multirow}
\usepackage{longtable, pbox, rotating, hvfloat}
\usepackage{booktabs}
\usepackage{breakcites}
\usepackage{multicol}
\usepackage{notoccite}

\usepackage[numbers]{natbib}
\usepackage{amsmath, amssymb}

\usepackage[table, dvipsnames]{xcolor}

\usepackage{tabularray}
\usepackage{epigraph}
\usepackage{fancyhdr}
\usepackage[colorlinks=true,citecolor=red,linkcolor=black]{hyperref}
\usepackage[Lenny]{fncychap}
\usepackage{scalerel}
\usepackage{pgfplots}
\usepackage{wrapfig}
\usepackage{caption}
\usepackage[notransparent]{svg}
\usepackage[margin=.9in]{geometry}
\usepackage{breakcites}
\usepackage{amsmath, amssymb}
\definecolor{lightgray}{gray}{0.983}

\usepackage{epigraph}
\usepackage{fancyhdr}
\usepackage[colorlinks=true,citecolor=red,linkcolor=black]{hyperref}
\usepackage[Lenny]{fncychap}
\usepackage{scalerel}
\usepackage{pgfplots}
\usepackage{wrapfig}
\usepackage{enumitem}
\usepackage{verbatim}
\usepackage{float}
\usepackage{graphicx , array}
\usepackage{adjustbox}

\usepackage{tikz}
\usepackage[graphicx]{realboxes}
\usepackage{pgfornament}

\usepgfplotslibrary{groupplots,dateplot}
\pgfplotsset{compat=newest}
\usetikzlibrary{positioning, patterns,calc, arrows}

\definecolor{navyblue}{rgb}{0.0, 0.0, 0.5}

\usetikzlibrary{positioning,arrows,calc}
\def\nodedist{35pt}
\def\layerdist{45pt}

\usepackage{subcaption}
\usepackage{makecell}
\usepackage{enumitem}
\usepackage{verbatim}
\usepackage{float}
\usepackage{graphicx}
\usepackage{adjustbox}

\usepackage{tikz}
\usepackage[graphicx]{realboxes}
\usepackage{pgfornament}

\usepgfplotslibrary{groupplots,dateplot}
\pgfplotsset{compat=newest}
\usetikzlibrary{positioning, patterns,calc, arrows}

\definecolor{navyblue}{rgb}{0.0, 0.0, 0.5}

\usetikzlibrary{positioning,arrows,calc}
\def\nodedist{35pt}
\def\layerdist{45pt}

\usepackage{bibentry}

\begin{document}

\begin{center}

	\rule{\linewidth}{3pt}
    \Large
	A Critical Review of Physics-Informed Machine Learning Applications in Subsurface Energy Systems
	\rule{\linewidth}{1pt}
\end{center}

Abdeldjalil Latrach\textsuperscript{1}, Mohamed L. Malki\textsuperscript{1, 2*}, Misael Morales\textsuperscript{2, 3}, Mohamed Mehana\textsuperscript{2}, Minou Rabiei\textsuperscript{1}

\textsuperscript{1} Energy and Petroleum Engineering Department, University of Wyoming\\
\textsuperscript{2} Environmental and Earth Sciences Group, Los Alamos National Lab\\
\textsuperscript{3} Petroleum and Geosystems Engineering, University of Texas at Austin\\
\textsuperscript{*} Corresponding author: \texttt{mlmalki@lanl.gov}

\leftskip=1cm
\rightskip=1cm

\paragraph{Abstract} Machine learning has emerged as a powerful tool in various fields, including computer vision, natural language processing, and speech recognition. It can unravel hidden patterns within large data sets and reveal unparalleled insights, revolutionizing many industries and disciplines. However, machine and deep learning models lack interpretability and limited domain-specific knowledge, especially in applications such as physics and engineering. Alternatively, physics-informed machine learning (PIML) techniques integrate physics principles into data-driven models. By combining deep learning with domain knowledge, PIML improves the generalization of the model, abidance by the governing physical laws, and interpretability. This paper comprehensively reviews PIML applications related to subsurface energy systems, mainly in the oil and gas industry. The review highlights the successful utilization of PIML for tasks such as seismic applications, reservoir simulation, hydrocarbons production forecasting, and intelligent decision-making in the exploration and production stages. Additionally, it demonstrates PIML’s capabilities to revolutionize the oil and gas industry and other emerging areas of interest, such as carbon and hydrogen storage; and geothermal systems by providing more accurate and reliable predictions for resource management and operational efficiency.


\leftskip=0pt\rightskip=0pt

\section{Introduction}
Artificial intelligence, particularly machine learning (ML) and deep learning (DL), has remarkably advanced over the past few decades \cite{LeCun2015}. These breakthroughs have been primarily fueled by the exponential growth of computational power and the abundance of big data, which can be effectively utilized for training and testing these models. Deep learning has revolutionized numerous scientific fields, ranging from computer vision and medical image diagnosis to natural language processing \Citep{Brown2020, Huang2020, Li2018}. Most of these models were developed in the broad deep-learning community to solve generic problems, where the availability of extensive labeled datasets, allowed training models with billions, and even trillions of parameters. On the other hand, the scientific community lacks this luxury of data abundance; additionally, scientists have more rigorous constraints imposed on their procedures and the associated outputs. Hence, we highlight several limitations that hinder the wide adoption of machine learning for serious and rigorous scientific research: 1) \textit{Lack of interpretability}: Despite the progress in improving neural networks interpretability, such as gradient-weighted class activation mapping (Grad-CAM) \Citep{Selvaraju2019} and attention maps \Citep{Hassanin2022}, ML models are still largely considered black boxes (or gray boxes at best). 2) \textit{Data requirements}: ML models required large datasets, usually unavailable in several scientific applications. 3) \textit{Abidance to physical laws}: ML models have no inherent and explicit tendency to respect physical laws, and their predictions can be physically nonsensical. 4) \textit{Extrapolation}: ML models are notoriously unable to extrapolate outside the distribution of training data. 

To address these limitations, there have been several emerging approaches to incorporate domain knowledge and physical constraints into ML models, an approach to which we will refer hereafter by physics-informed machine learning (PIML). In the context of PIML, the integration of theoretical laws acts as \textit{informative prior}  to the model and introduces \textit{inductive biases} that steer the training process in the direction of the most physically-plausible model \cite{Battaglia2018} by constraining the space of possible solutions \cite{Karniadakis2021}. This is especially useful for small data regimes where high-quality data are scarce or unavailable.

Using neural network to solve physical systems governed by ordinary, partial, and stochastic differential equations (ODEs, PDEs, and SDEs) generated great interest in PIML research and will make up the bulk of this study. This process is termed \textit{neural simulation}. We can distinguish two types of neural simulators: 1) \textit{neural solvers} which aim to solve ODEs, PDEs, or SDEs using neural networks, and 2) \textit{neural operators} which aim to learn solution maps of parametric differential equations \cite{Hao2023}. First, it is important to understand how neural simulators compare to traditional numerical solvers such as finite difference and finite element methods (FDM and FEM), since both approaches aim to solve the same problem. This comparison is presented in Table \ref{table:comparison}.

\begin{longtblr}[
  label = table:comparison,
  caption = {Comparison between numerical and neural simulators for modeling real-word systems},
  entry = none,
]{
  width = \linewidth,
  colspec = {Q[3]Q[10]Q[10]},
  hlines = {0.03em},
  hline{1,2, 9} = {-}{0.1em},
}
                   & Numerical simulators                                                                                                            & Neural simulators                                                                                                                                     \\
Principle          & Discretize the governing equations and approximate the solution by a set of algebraic     equations that are numerically solved & Directly approximate the governing equations by optimizing a set of learnable parameters                                                                \\
Meshing            & Mesh-based                                                                                                                      & Meshless                                                                                                                                              \\
Accuracy           & Very reliable and high accuracy (given sufficiently fine meshing)                                                               & Lower accuracy compared to numerical simulators                                                                   \\
Computation        & Computationally expensive especially for fine meshes, which are required to achieve     accurate solutions                      & Training is computationally expensive, but inference is near real-time                                                                                \\
Incomplete physics & Complete physical formulation is a must                                                                                         & Can handle ill-posed problems with     incomplete physical understanding                                                                              \\
Data incorporation & Incorporating sensors and experimental data can be challenging                                                                  & Efficiently     incorporates data from various sources                                                                                                \\
Artifacts          & Can suffer from certain artifacts such as numerical dispersion and diffusion                                                    & Can suffer from handling shocks in the solution and Multiphysics problems
\end{longtblr}

Numerical simulators possess a strong and mature theoretical foundation. There is no intent---for the moment---to replace them with neural simulators for critical applications that require highly accurate and robust solutions. However, as we shall see later, neural simulators can complement numerical simulators, accelerate simulation workflows, support the findings, and enhance decision-making. 

This study represents a comprehensive and critical review of state-of-the-art implementations of PIML in subsurface energy systems applications, with emphasis on the oil and gas industry. We organized this paper as follows: Section \ref{section:modes} explains the various modes of integrating physics into data-driven models. Section \ref{section:theory} introduces one of the most common PIML models, the physics-informed neural networks (PINNs). Section \ref{section:applications} emphasizes the current state-of-the-art PIML implementations in the various areas of subsurface engineering, namely: geoscience, drilling engineering, reservoir engineering, and production forecasting and optimization. Section \ref{section:improvement} presents the best-practices for successful implementation of the PIML framework to solve real-word issues, and Section \ref{section:conclusion} highlights the knowledge gaps and discusses potential future areas of research.

\section{Modes of integrating physics into data-driven models}
\label{section:modes}

\begin{figure}[h]
    \centering
	\def\itemdist{.55in}
\begin{tikzpicture}[approach/.style={align=left, text width=2.15in, rectangle,rounded corners=2pt, minimum width = 2.3in, draw, color = white!40!black, minimum height = .5in}, example/.style={fill = white!95!black, minimum width=1.3in, minimum height=0.15in},
	bars/.style={shading = axis,rectangle,color=white, left color={rgb:red,255;green,26;blue,18},rounded corners=2.5pt,  right color={rgb:red,2;green,95;blue,180}, minimum width=6*\itemdist-.05in,rounded corners=2.5pt, minimum height=0.3in, rotate=-90 }]
	\node[approach] at (0, 0) (data) {Data};
	\node[approach] at (0, -\itemdist) {\footnotesize Postprocessing};
	\node[approach] at (0, -2*\itemdist) {\footnotesize Initialization};
	\node[approach] at (0, -3*\itemdist) {\footnotesize Optimizer\\design};
	\node[approach] at (0, -4*\itemdist) {\footnotesize Architecture\\design};
	\node[approach] at (0, -5*\itemdist) {\footnotesize Loss\\function};
    \node [bars ,shading angle=180] (box1) at (1.35in, -2.5*\itemdist){\scriptsize Physical knowledge};
    \node [bars ,shading angle=0] (box2) at (1.7in, -2.5*\itemdist){\scriptsize Data availability};
	\node[rotate=-90, color=white] at  (1.35in, -2.5*\itemdist+1.4in){\scriptsize less};
	\node[rotate=-90, color=white] at  (1.35in, -2.5*\itemdist-1.4in){\scriptsize more};
	\node[rotate=-90, color=white] at  (1.7in, -2.5*\itemdist+1.4in){\scriptsize more};
	\node[rotate=-90, color=white] at  (1.7in, -2.5*\itemdist-1.4in){\scriptsize less};
	\node[approach, rotate=-90, align=center, text width=5*\itemdist, minimum width=6*\itemdist-0.        05in] at (-1.45in, -2.5*\itemdist) {\footnotesize Hybrid approaches};

	\node[example] at (0.4in, 0.1in) {\scriptsize Synthetic data};
	\node[example] at (0.4in, -0.1in) {\scriptsize Experimental data};

	\node[example] at (0.4in, -\itemdist+0.1in) {\scriptsize Continuity constraint};
	\node[example] at (0.4in, -\itemdist-0.1in) {\scriptsize Smoothness constraint};

	\node[example] at (0.4in, -2*\itemdist) {\scriptsize Transfer learning};

	\node[example] at (0.4in, -3*\itemdist) {\scriptsize Physics awareness};

	\node[example] at (0.4in, -4*\itemdist+0.1in) {\scriptsize Invariances};
	\node[example] at (0.4in, -4*\itemdist-0.1in) {\scriptsize Symmetries};

	\node[example] at (0.4in, -5*\itemdist+0.1in) {\scriptsize ODEs/PDEs/SDEs};
	\node[example] at (0.4in, -5*\itemdist-0.1in) {\scriptsize Conservation laws};
\end{tikzpicture}
    \caption{Various modes of physics integration within data-driven models, some examples for each mode, and where they fall on the physical knowledge/data availability spectrum.}
    \label{fig:approaches}
\end{figure}
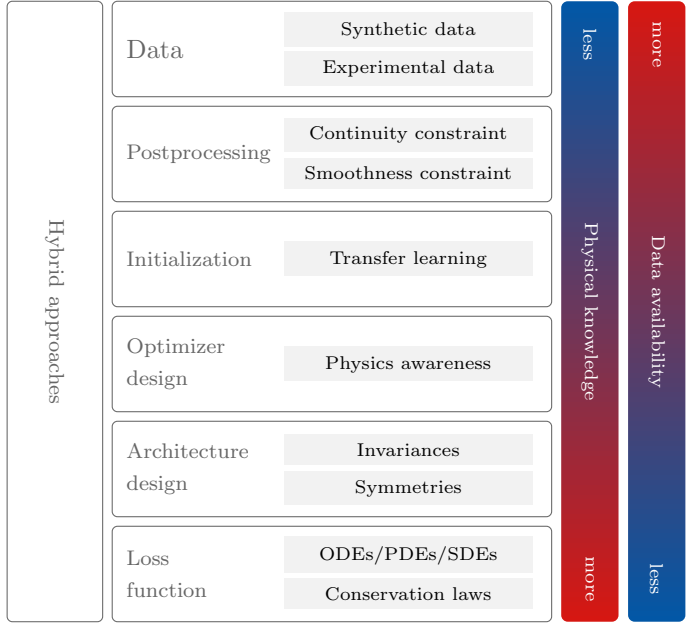

Throughout the body of literature examined in this study, the terms physics-informed, physics-guided, or physics-constrained data-driven, machine learning or deep learning are used to denote several approaches that we gathered under the umbrella term of PIML. Despite the nuanced distinction, all of them  share a common objective: integrating physical knowledge into data-driven models. In this section, we elucidate several modes of integration, namely: 1) data and feature engineering, 2) postprocessing, 3) initialization, 4) optimizer design, 5) architecture design, 6) loss function, and 7) hybrid models. The selection of a particular mode of physics integration is not arbitrary, but rather contingent upon two key factors: 1) the extent of physical understanding of the problem to be solved and 2) the availability of relevant data. Nonetheless, strict guidelines are not prescriptive, and various approaches can be employed to tackle the same problem or even be synergistically combined to yield novel hybrid models. Figure \ref{fig:approaches} represents the different physics integration modes and where they fall on the data availability/physical knowledge spectrum. 

\paragraph{Data and feature engineering}
The most rudimentary form of physics integration involves utilizing physics and engineering data to construct a conventional neural network. This methodology relies on the universal function approximation theory, which states that neural networks can approximate any continuous function (using one hidden layer) using a sufficiently large network \cite{Hornik1989}. However, this represents the most limited form of incorporating physical priors into a data-driven model. It possesses distinct drawbacks, namely: 1) lack of transparency, as the model's inner workings and the insights it gains regarding correlations and patterns remain obscure, relying solely on the function approximation capabilities of neural networks, and 2) the demand for copious amounts of data, given that these vanilla deep neural network architectures are characterized by a large number of parameters which require substantial data quantities for proper training and convergence.

\paragraph{Postprocessing}
The prediction generated by a data-driven model, such as a neural network, can undergo postprocessing to enforce specific physical properties on the final predictions. Postprocessing techniques serve as a means to ensure that the model's outputs adhere to fundamental physical principles, which is crucial when dealing with rigorous scientific applications. For instance, certain postprocessing methods can be employed to enforce continuity or smoothness in the final predictions, aligning them with the expected behavior of physical systems (e.g., ensuring spatial continuity in a predicted pressure field). By integrating postprocessing into the modeling pipeline, we can enhance the overall robustness and validity of the predictions. Such postprocessing acts as a filter that can identify and rectify unsound predictions that might violate physical laws or constraints. While postprocessing represents a relatively weak form of physics integration compared to fully physics-based models, it complements the data-driven approach by refining the output and bringing it closer to the expected physical behavior.

\paragraph{Initialization}
Proper initialization of parameters in ML models holds utmost importance during the training process. The choice of initial weights and biases significantly influences the model's convergence \Citep{Sutskever2013, Willard2020}. On the contrary, a poorly initialized model may struggle to converge effectively, leading to be trapped in low-quality local minima. \textit{Transfer learning} is an ML technique whereby a neural network is \textit{pretrained} (i.e., source learner) to solve a generic, easier problem  (i.e., source domain), and then it is \textit{fine-tuned} (i.e., target learner) on a more specialized, relatively harder-to-solve problem (i.e., target domain). There are various parameter control strategies for transfer learning, namely: 1) \textit{parameter sharing} where most of the weights from the target learner are frozen, and the last few layers are fine-tuned in the target domain; and 2) \textit{parameter restriction} where the target learner's parameters are restricted to be similar to those of the source learner \Citep{Zhuang2020}. This approach offers several advantages: firstly, it leverages the knowledge captured by pretrained models, leading to improved performance. Secondly, it reduces data requirements in the target domain, which is particularly beneficial for real-world scientific and engineering problems where data might be limited. Lastly, it cuts down on computation costs, as the same pretrained source model can be reused for several tasks and just needs to be fine-tuned on the target dataset of interest. As an example in the context of PIML, models can be pretrained on more generic problems or abundant simulation data, and then fine-tuned on specific problems where experimental data is scarce.

\paragraph{Optimizer design}%
\label{sub:optimizer_design}
Optimizers used to minimize the network's loss function and learn the parameters can be constructed to be physically-aware. To the best of our knowledge, this is the least researched area within PIML literature, and is only included eere for the sake of exhaustivity, since there exist examples for this approach in the broad literature \cite{Yi2022}. This can be an area of potential research, although designing optimizers is less straightforward than other modes of integration.

\paragraph{Architecture design}
The model's architecture can be leveraged to encode certain physical attributes or characteristics of the problem of interest. The traditional fully connected neural network (FCNN) cannot accomplish this by default, yet it can be modified in certain ways to do that, such as fixing certain parameters of the model to physically meaningful values \Citep{Sun2020} or introducing intermediate physical variables \Citep{Daw2019, Willard2020}. Other more sophisticated architectures can encode certain invariances and symmetries, thus constraining the space of possible solutions. The most prominent example of such architectures is convolutional neural networks (CNN) which consist of convolutional layers that can encode translation and rotation invariances. Graph neural networks (GNN) operate on graph-structured data which is more representative of many real-world systems (e.g., wells' spatial distribution in an oilfield) than the simple FCNN architecture. Additionally, tensor basis neural networks (TBNN) have emerged as a powerful tool for tackling fluid flow and turbulence problems. TBNNs utilize an invariant tensor basis, such as the Reynolds stress anisotropy tensor, to embed physical invariance properties into the deep neural network architecture \Citep{Ling2016}. By leveraging these tensor basis neural networks, researchers have achieved notable improvements in predicting Reynolds stress anisotropy and other complex fluid dynamics phenomena.

\paragraph{Loss function}
Data-driven models often fail to capture dynamics of complex systems with many interacting physical variables in both spatial and temporal domains, which is especially true for small-data regimes where data scarcity exacerbates the issue. The aforementioned modes impose certain informative biases on the training process, yet the most explicit form of physics integration is to directly encode the physical laws (e.g., ODEs/PDEs/SDEs and conservation laws) into the loss function of the model. This will ensure that the resulting model will produce predictions that align the most with the underlying physics. 

PINN is a variation of the vanilla fully connected architecture that leverages automatic differentiation to impose the governing PDEs into the loss function \Citep{Raissi2019}. PINNs can be used for both solving and discovering governing equations. In the first task, or forward modeling problem, the network is trained to minimize the residual form of the differential equation, producing physically consistent predictions. For the second task, or inverse modeling problem, the parameters of the governing equation are set to be learnable parameter, which are determined during the training process. This approach exhibits several advantages: 1) explicit encoding of physical laws results in strong form of regularization that steers the model into the direction of physically consistent results, 2) allows for training without unlabeled data since collocation points can be randomly or systematically sampled from the domain of interest and used for training, and 3) the model may even extrapolate beyond the range of training data.

\paragraph{Hybrid models}
Several physics integration modes can be employed together to achieve a hybrid model. This approach depends on several factors such as the degree of physical knowledge and available data. For example, a certain neural architecture that imposes its own constraints can also be coupled with a physics-based loss function for further regularization and enforcement of physical laws. \citet{WendiLiu2023} used a graph neural network to represent an oil field under waterflood settings where the various wells demonstrate subsurface connectivity, then combined it with a physics-based (i.e., material balance equation) loss function. Graphs are a better representation of the spatial distribution of oil wells and can better encode the wells connectivity in a more intuitive manner, while the physics-based loss function further enforces the underlying physics.

\section{Theoretical formulation of physics-informed neural networks}
\label{section:theory}
	In this section, we provide a theoretical introduction to PINNs defined by \citet{Raissi2019} for both PDEs approximation and discovery. We will discuss the mathematical formulation of PDEs, initial and boundary conditions, and their integration within neural networks.

\subsection{Partial differential equations, initial, and boundary conditions}
PDEs are mathematical equations that involve functions and their partial derivatives. They are used to describe a wide range of phenomena in physics, engineering, and other scientific fields, including heat transfer, fluid flow, and wave propagation. PDEs allow us to study how these quantities change with respect to multiple variables and their interactions. Solving PDEs can be challenging due to their complex mathematical properties, such as nonlinearity and the need to satisfy initial and boundary conditions. Analytical solutions are often limited to simple cases, leading to the development of numerical methods that approximate the solutions numerically.

Physics-informed neural networks emerged as a data-driven, mesh-free approach to solve PDEs. 

Consider the general form of a partial differential equation given by:

\begin{equation}
    \mathcal{N}[u(x, t)] = 0, \quad x \in \Omega \subset \mathbb{R}^d ,\quad t \in [0, T]
    \label{eq:pde}
\end{equation}

where $u(x, t)$ is the unknown function (i.e., hidden solution) and $\mathcal{N}[\cdot]$ is a differential operator. The differential operator $\mathcal{N}$ can be classified into three categories: 1) parabolic (e.g., diffusion equation), 2) hyperbolic (e.g., wave equation), or 3) elliptic (e.g., Laplace's equation). $x$ and $t$ are variables describing the location in space and time, and $\Omega$ and $T$ describe the spatial and temporal domains. Besides the need to satisfy the PDE described by the differential operator, the solution also needs to satisfy given initial and boundary conditions, to ensure the uniqueness of the solution. Initial conditions are values of $u(x, t)$ given for $t=t_0$ and boundary conditions are the values of $u(x, t)$ at the boundaries of the domain at all times. Boundary conditions can be Dirichlet, Neumann, Robin, or mixed boundary conditions.

\begin{itemize}
    \item \textbf{Dirichlet b.c.}: specified fixed values for the unknown function at the boundaries of the domain.
    \begin{equation}
        u(x, t) = f(x), \quad \forall x \in \partial \Omega
    \end{equation}
    where $\partial\Omega$ is the boundary of the domain of interest and $f$ is a scalar function defined on this boundary.
    \item \textbf{Neumann b.c.}: Specify the derivative of the unknown function $u$ at the boundaries of the domain.
    \begin{equation}
        \frac{\partial u(x, t)}{\partial \eta} = f(x), \quad \forall x \in \partial \Omega
    \end{equation}
    where $n$ is the outward normal to the boundary of the domain.
    \item \textbf{Robin b.c.}: Specify the boundary conditions as a combination of Dirichlet and Neumann b.c. Mathematically, can be written as:
    \begin{equation}
        au(x, t) + b\frac{\partial u(x, t)}{\partial \eta} = f(x), \quad \forall x \in \partial \Omega
    \end{equation}
    where $a$ and $b$ are non-zero constants.

    \item \textbf{Mixed b.c.}: In this scenario, different types of boundary conditions are used in different parts of the domain
\end{itemize}

\subsection{PINNs for solving governing equations}
The unknown solution $u(x, t)$ to the PDE is approximated using a deep neural network such as $\hat{u}_\theta(x, t) \approx u(x, t)$ where $\hat{u}_\theta$ is the function learned by the neural network and parametrized by a set of parameters $\theta$. The parameters $\theta$ are the weights and biases of the neural network:

\begin{equation}
    \theta = \left\{W_1, W_2, \dots, W_n, b_1, b_2, \cdots, b_n\right\}
\end{equation}

and these parameters can be learned by solving an optimization problem of the form:

\begin{equation}
    \underset{\theta}{\text{argmin}} \quad\mathcal{L}_{\textrm{\tiny IB}}(\theta) +  \mathcal{L}_{\textrm{\tiny PDE}}(\theta) + \mathcal{L}_{\textrm{\tiny Data}}(\theta)
    \label{eq:loss}
\end{equation}

where $\mathcal{L}_{\textrm{\tiny IB}}(\theta)$, $\mathcal{L}_{\textrm{\tiny PDE}}(\theta)$, and $\mathcal{L}_{\textrm{\tiny Data}}(\theta)$ are the losses associated with the function $\hat{u}_\theta(x, t)$ when evaluated at initial and boundary conditions, on the collocation points, and on available real-world data, measured as the mean squared error. Equation \ref{eq:loss} can therefore be also written more explicitly as:

\begin{align}
    \underset{\theta}{\text{argmin}} \quad
	\frac{1}{N_{\textrm{\tiny IB}}} &\sum_{i=1}^{N_{\textrm{\tiny IB}}}\left[\hat{u}_\theta(x^i_{\textrm{\tiny IB}}, t^i_{\textrm{\tiny IB}})-u^i_{\textrm{\tiny IB}}\right]^2 + \\ \nonumber 
	\frac{1}{N_{\textrm{\tiny PDE}}}  & \sum_{i=1}^{N_{\textrm{\tiny PDE}}}
	\mathcal{N}\left[\hat{u}_\theta(x^i_{\textrm{\tiny PDE}},t^i_{\textrm{\tiny PDE}} )\right]^2 +\\ \nonumber 
	 \frac{1}{N_{\textrm{\tiny Data}}} & \sum_{i=1}^{N_{\textrm{\tiny Data}}}\left[\hat{u}_\theta(x^i_{\textrm{\tiny Data}}, t^i_{\textrm{\tiny Data}})-u^i_{\textrm{\tiny Data}}\right]^2
\end{align}

where $\{x^i_{\textrm{\tiny IB}}, t^i_{\textrm{\tiny IB}}, u^i_{\textrm{\tiny IB}} \}_{i=1}^{N_{\textrm{\tiny IB}}}$ and $\{x^i_{\textrm{\tiny Data}}, t^i_{\textrm{\tiny Data}}, u^i_{\textrm{\tiny Data}} \}_{i=1}^{N_{\textrm{\tiny Data}}}$ are the initial/boundary condition and real-world measurements data points, respectively, while $\mathcal{N}\left[\hat{u}_\theta(x^i_{\textrm{\tiny PDE}},t^i_{\textrm{\tiny PDE}} )\right]$ is the differential operator applied to the approximated function evaluated on chosen collocation points $\{x^i_{\textrm{\tiny PDE}}, t^i_{\textrm{\tiny PDE}} \}_{i=1}^{N_{\textrm{\tiny PDE}}}$.

Sometimes, the data loss term $\mathcal{L}_{\textrm{\tiny Data}}(\theta)$ is dropped if real-world data is not available or is of low quality, and only the initial/boundary conditions and PDE residual terms are left, which are crucial to solve the PDE of interest. Furthermore, scaling coefficients can be introduced in Equation \ref{eq:loss} to control the contribution of each term to the total loss. Equation \ref{eq:loss} can therefore be written as:

\begin{equation}
    \underset{\theta}{\text{argmin}} \quad \lambda_{\textrm{\tiny IB}} \mathcal{L}_{\textrm{\tiny IB}}(\theta) +  \lambda_{\textrm{\tiny PDE}}\mathcal{L}_{\textrm{\tiny PDE}}(\theta) + \lambda_{\textrm{\tiny Data}} \mathcal{L}_{\textrm{\tiny Data}}(\theta)
    \label{eq:loss_regularized}
\end{equation}

which can significantly alter the training process. This is discussed further in the context of curriculum learning in Section \ref{ssub:curriculum}.

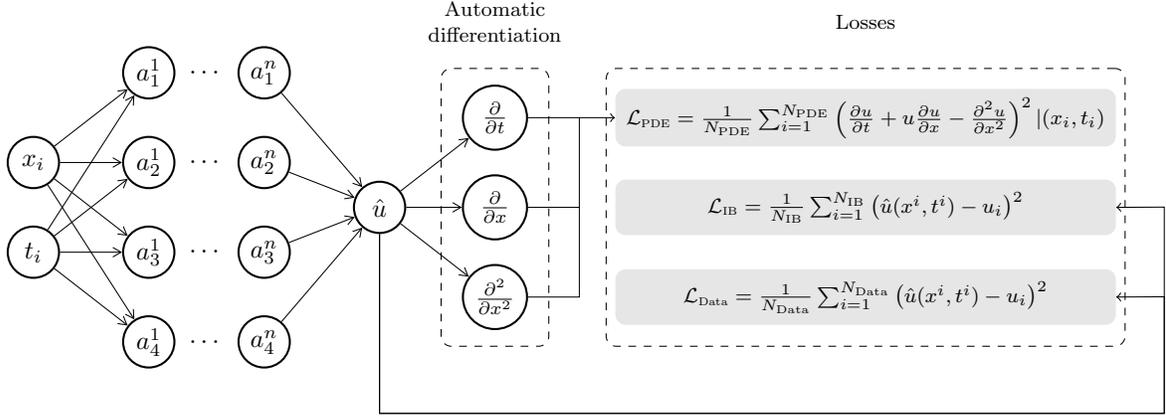
\begin{figure}[h]
    \centering
	\scalebox{.97}{%
\begin{tikzpicture}[neuron/.style={circle,draw, thick,align=center, minimum size=2em,inner sep=1pt}, loss/.style={rectangle, minimum width = 2.7in, rounded corners, fill = white!90!black, minimum height = 0.3in},derivative/.style={circle,draw, thick,align=center, minimum size=2.5em,inner sep=1pt}, input/.style={->}]
    




\foreach \y in {1,...,2} \node[neuron]  (I\y) at (0,-\y*\nodedist) {};  
\node[text width=2cm, align=center] at (0,-1*\nodedist) {$x_i$};
\node[text width=2cm, align=center] at (0,-2*\nodedist) {$t_i$};


\foreach \y in {1,...,4} \node[neuron]  (H1\y) at ($(\layerdist,-\y*\nodedist) +(0, 1*\nodedist)$) {$a_\y^1$};
\foreach \y in {1,...,4} \node[]  (dots\y) at ($(1.5*\layerdist,-\y*\nodedist) +(0, 1*\nodedist)$) {$\cdots$};

\foreach \y in {1,...,4} \node[neuron]  (H2\y) at ($(2*\layerdist,-\y*\nodedist) +(0, 1*\nodedist)$) {$a_\y^n$};

\foreach \y in {1,...,1} \node[neuron] (O\y) at ($(I2) + (3*\layerdist, 0.5*\nodedist)$) {$\hat{u}$};

\node[derivative] (D1) at ($(O1) + (\layerdist, \nodedist)$) {$\frac{\partial}{\partial t}$};
\node[derivative] (D2) at ($(O1) + (\layerdist, 0)$) {$\frac{\partial}{\partial x}$};
\node[derivative] (D3) at ($(O1) + (\layerdist, -\nodedist)$) {$\frac{\partial^2}{\partial x^2}$};
\node[text width=4cm, align=center] (r) at ($(D2) + (0, 1in)$) {\footnotesize \baselineskip=10pt Automatic\\differentiation\par};
\node[rectangle, draw, dashed, minimum width = 1.2*\nodedist, rounded corners,  minimum height = 1.5in] (autodiff) at ($(D2)$) {};

\node[loss] (loss1) at ($(D1) + (2in, 0)$) {\footnotesize$\mathcal{L}_{\scaleto{\textrm{PDE}}{3pt}} = \frac{1}{N_{\scaleto{\textrm{PDE}}{3pt}}}\sum_{i=1}^{N_{\scaleto{\textrm{PDE}}{3pt}}}\left(\frac{\partial u}{\partial t} + u\frac{\partial u}{\partial x}-\frac{\partial^2 u}{\partial x^2}\right)^2 | (x_i, t_i)$};
\node[loss] (loss2) at ($(D2) + (2in, 0)$) {\footnotesize$\mathcal{L}_{\scaleto{\textrm{IB}}{3pt}} = \frac{1}{N_{\scaleto{\textrm{IB}}{3pt}}} \sum_{i=1}^{N_{\scaleto{\textrm{IB}}{3pt}}}\left(\hat{u}(x^i, t^i)-u_i\right)^2$};
\node[loss] (loss3) at ($(D3) + (2in, 0)$) {\footnotesize$\mathcal{L}_{\scaleto{\textrm{Data}}{3pt}} = \frac{1}{N_{\scaleto{\textrm{Data}}{3pt}}} \sum_{i=1}^{N_{\scaleto{\textrm{Data}}{3pt}}}\left(\hat{u}(x^i, t^i)-u_i\right)^2$};

\draw[->] (O1.south) -- ++(0,-2*\nodedist)  -|  ++(6.8*\layerdist,2*\nodedist+1.05em) --  (loss2.east);	
\draw[->] (O1.south) -- ++(0,-2*\nodedist)  -|  ++(6.8*\layerdist,1*\nodedist+1.05em) --  (loss3.east);

\node[rectangle, draw, dashed, minimum width = 2.8in, rounded corners,  minimum height = 1.5in] (autodiff) at ($(loss2)$) {};
\node[text width=4cm, align=center] (r) at ($(loss2) + (0, 1in)$) {\footnotesize \baselineskip=10pt Losses\par};





\draw[->] (D1.east) --  (loss1.west);	
\draw[] (D2.east) -- ++(1em,0)  -|  ++(1em,\nodedist)  ;	
\draw[] (D3.east) -- ++(1em,0)  -|  ++(1em,\nodedist)  ;	
\foreach \dest in {1,...,4} \foreach \source in {1,...,2} \draw[->, >=angle 60] (I\source) -- (H1\dest);

\foreach \dest in {1,...,1} \foreach \source in {1,...,4} \draw[->, >=angle 60] (H2\source) -- (O\dest);

\foreach \dest in {1,...,3} \foreach \source in {1,...,1} \draw[->, >=angle 60] (O\source) -- (D\dest);


\end{tikzpicture}
		
}
    \caption{Typical architecture of a physics-informed neural network}
    \label{fig:architecture}
\end{figure}

Automatic differentiation is used for calculating the different terms in the PDE defined by the operator $\mathcal{N}[\cdot]$, which are combined to form the PDE residual. Figure \ref{fig:architecture} shows a typical architecture of a physics-informed neural network with loss terms accounting for the PDE residual, initial/boundary conditions, and data mismatch.

\subsection{PINNs for discovering governing equations}%
\label{sub:pinns_for_partial_differential_equations_discovery}

As opposed to solving known governing equations, the PINN framework also be for discovery of these equations from scarce and noisy data. Let's take the residual form of the acoustic wave equation for example, which is used in various seismic applications:

\begin{equation}
	\frac{\partial^2 p}{\partial x^2} - \frac{1}{c^2}\frac{\partial^2p}{\partial t^2} = 0
	\label{eq:acoustic}
\end{equation}

where $p$ is the acoustic pressure representing the state variable, and $c$ is the speed of sound in the medium. We can rewrite Equation \ref{eq:acoustic} assuming the various coefficients in the equations to be unknown:

\begin{equation}
	\lambda_1\frac{\partial^2 p}{\partial x^2} + \lambda_2\frac{\partial^2p}{\partial t^2} = 0
\end{equation}

The same process for solving differential equations holds true, the only difference is that $\lambda_1$ and $\lambda_2$ are set to be learnable parameters and will be determined through the learning process rather than being explicitly specified. Ideally, after the training process, these parameters will converge to $\lambda_1\approx 1$ and $\lambda_2\approx -\frac{1}{c^2}$. The starting point for this problem is the mere assumption that the governing equations is a certain combination of the second derivatives of pressure with respect to time and space.  This approach can be generalized for any physical problem where we can speculate which variables may affect the system's dynamics, and then determine the interplay of these variables. However, discovery of governing equations requires labeled data for the determination of the equation's parameters. Nevertheless, it can work reasonably well with datasets of limited size even with the presence of noise \Citep{Raissi2019}.


\section{Machine learning in subsurface energy systems: overview and the shift towards PIML}
\label{section:applications}

Since their inception, neural networks have been employed to solve almost every conceivable problem in oil and gas \citep{Alkinani2019}, spanning various domains, such as 
exploration \citep{Hansen1993, Karrenbach2000, Huang2006, Verma2012, Ross2017}, drilling 
\citep{Arehart1990, Dashevskiy1999, Fruhwirth2006, Moran2010, Elkatatny2016}, production \citep{Thomas1995, Shelley1999, Ghahfarokhi2018, Pankaj2018, Dogail2018} and reservoir characterization and simulation \citep{An1993, Ayoub2007, Elshafei2009, Ma2015}. The architectural complexity of the employed model drastically increased during the past few decades, from simple feed-forward artificial neural networks, convolutional, recurrent and graph neural networks, to more advanced, even state-of-the-art approaches such as transformer-based architectures.

Despite the progress that has been made, physics integration within data-driven models is still lacking, where it mostly consists of the first and, to a lesser degree, second modes of integration (i.e., data and architecture design).
Nevertheless, recently there has been a shift in adopting the PIML approach  in the oil and gas industry across its areas, which is reflected in the growing body of literature. The currently available literature can be categorized into five main distinct categories: 1) geology and geophysics, 2) drilling engineering, 3) reservoir engineering, 4) production forecasting and optimization, and 5) CO\textsubscript{2} storage. 


\begin{longtblr}[
  label = table:summary,
  caption = {Summary of modes of various modes of physics integration for each area of study},
  entry = none,
]{
  width = \linewidth,
  colspec = {Q[15]Q[13]Q[10]Q[10]Q[10]Q[10]},
  hline{1,2, 10} = {1-6}{0.1em},
  hline{2-9} = {1-6}{0.03em},
}
	 &  Data \& feature engineering & Transfer\newline learning & Architecture  design &Loss function & Postprocessing \\
Geoscience&  \cite{Song2020} \cite{Alkhalifah2020} \cite{binWaheed2020}   &                   &\cite{McAliley2021} \cite{binWaheed2022} \cite{Cheng2022} \Citep{Wei2022} \Citep{Li2023}\Citep{Yang2021}\Citep{Dhara2022}\Citep{Huang2022b}& \Citep{Huang2022b}\cite{Alkhalifah2020b} \cite{Song2020} \cite{Song2021} \cite{Gou2023} \cite{Voytan2020} \cite{Moseley2020} \cite{Kumar2020} \cite{Alkhalifah2021} \cite{Huang2022} \cite{YijieZhang2023}\Citep{Xu2019}\Citep{binWaheed2021}	&  \\
	Drilling & \cite{Kharazmi2021}\cite{Kaneko2023}& \cite{Tang2022}                   &   \cite{Kharazmi2021}    \cite{Prasham2022}        & \cite{Tang2022} &        \\
	Reservoir characterization \& simulation		&\Citep{Zhang2022} \Citep{Lv2021} \cite{Yoga2022}  \cite{Zhang2021}  &     \cite{Hadjisotiriou2023}              &\Citep{Grttner2023} \Citep{Boateng2017} \cite{Zhang2021}    \cite{Tembely2021} \cite{Wu2018}          &    \cite{Ihunde2021} \cite{Behl2023}\Citep{Wu2023} \Citep{Deng2021} \Citep{Abbasi2023} & \cite{Wang2022}\Citep{Chen2023} \\
	Fluid flow	&      & &\Citep{Haghighat2022}\Citep{Diab2022} \Citep{Torrado2021} \Citep{Diab2023} \Citep{Maucec2022} \cite{Battaglia2016} \Citep{KaiZhang2022} \Citep{Wen2022}      \cite{ZhaoZhang2022} &  \cite{Fuks2020} \cite{Magzymov2021} \cite{Gasmi2021} \cite{Alhubail2022}\cite{ZhaoZhang2022} \Citep{Sambo2021} \Citep{Shen2022}\Citep{Almajid2020} \Citep{Torrado2021}\Citep{Coutinho2023}\Citep{Haghighat2022} \Citep{He2020}&        \\
	Waterflooding \& EOR&  \Citep{Manasipov2023}   \Citep{Nagao2023} &                   & \Citep{Liu2023} \Citep{Darabi2022} \Citep{WendiLiu2023}& \Citep{Sarma2017} \Citep{Darabi2022} \Citep{WendiLiu2023} \Citep{Gladchenko2023} \Citep{Behl2023}  \Citep{Maniglio2021}      \Citep{Nagao2023}       &        \\
	Production forecasting	& \Citep{Molinari2021} \Citep{Harp2021} \cite{Manasipov2023} \cite{Staff2020} \cite{Razak2021c}   &  \cite{Razak2022}                  &    \cite{Razak2022}        &   \Citep{Molinari2021} \Citep{Busby2020} \Citep{Franklin2022} &\\
CO\textsubscript{2} storage &      &                   & \Citep{MingliangLiu2023} \Citep{Tariq2023} \Citep{Jiang2023} \Citep{Yan2022}& \Citep{HonghuiDu2023} \Citep{Shokouhi2021} \Citep{MingliangLiu2023} \Citep{Yan2022b} &     \Citep{Yan2022} \\
\end{longtblr}

\newcommand{\STAB}[1]{\begin{tabular}{@{}c@{}}#1\end{tabular}}

\subsection{Geology and geophysics}
PIML is a powerful tool in geoscience, offering innovative solutions for solving complex problems related to wave propagation, seismic inversion, and subsurface characterization. Here, we discuss the multitude of applications ranging from solving governing equations to replacing full seismic inversion workflows using state-of-the-art models.

\subsubsection{Modeling wave propagation in the subsurface: solving the governing equations.}
Accurately modeling seismic wave propagation is crucial for seismic imaging workflows and subsurface characterization. The \textit{acoustic wave equation} is the fundamental governing equation for wave propagation in a material medium (e.g., earth subsurface). The equation is a second-order hyperbolic PDE that relates the second derivative of pressure with respect to space coordinates, to its second time derivative. Solving the acoustic wave equation allows for understanding of wave propagation, its interaction with boundaries and various phenomena such as reflection, refraction and diffraction. The time-independent formulation of the wave equation in the frequency domain is known as the \textit{Helmholtz equation}. In high-frequency regimes, the \textit{eikonal equation} is an approximation to the Helmholtz equation whose solution provides valuable information about travel times of wavefronts.

The focus of most of PIML applications in the surveyed literature within geoscience applications is to solve these governing equations. The most straightforward implementation is imposing the second-order wave equation and the associated boundary conditions in the loss function of a PINN to obtain velocity models \Citep{Moseley2020, Voytan2020, Kumar2020, Xu2019} or first-order wave equation to estimate both velocity and density fields \Citep{YijieZhang2023}.

\begin{figure}[htpb]
	\centering
	\includegraphics[width=0.6\linewidth]{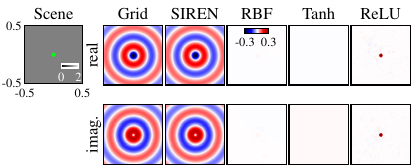}
	\caption{Helmholtz equation solution for a single point source (green dot) with uniform wave propagation velocity. The sinusoidal representation networks (SIREN) provide the most accurate solution when compared against grid solver \cite{Sitzmann2020}.}%
	\label{fig:siren}
\end{figure}

While solving the wave equation in the time domain has been the conventional approach in seismic applications, frequency domain solutions (i.e., the Helmholtz equation) provide several advantages (e.g., dimensionality reduction); this has gained prominence, particularly with the advancement of waveform inversion techniques (e.g., full-waveform inversion). PINNs were applied extensively to solve wave propagation in the subsurface under different settings with promising results \Citep{Alkhalifah2021,Alkhalifah2020b,Song2020}. It is noteworthy that replacing the traditional inverse tangent and other traditional activation functions with adaptive sinusoidal activation functions proved to result in a better convergence of the network by allowing it to learn more complex functions, as shown in Figure \ref{fig:siren} \Citep{Song2021,Sitzmann2020}. The effect of activation functions on network convergence will be discussed in greater detail later. In an alternative approach, PINNs can be used to learn Green's functions, a mathematical tool for solving PDEs, that satisfy the wave equation \Citep{Alkhalifah2020}. In addition to predicting wavefield solutions, variations of this architecture can locate subsurface seismic sources with a high precision \Citep{Grubas2021, Huang2022}.

Although travel times of seismic waves, obtained by solving the eikonal equation, provide less information than full seismic records, they are a crucial piece of information for various seismic applications. Similar to the aforementioned works, PINNs were also used to solve the eikonal equation, especially the anisotropic form of it, which is computationally expensive for traditional numerical solvers \Citep{binWaheed2020}. The PINN approach can even provide higher accuracy than conventional methods, such as finite-difference solutions that sometimes can rely on ill-posed physics of wave propagation \Citep{binWaheed2021} and can also be extended with Bayesian statistics for uncertainty quantification \Citep{Gou2023}. 

\begin{figure}[h]
	\centering
	\includegraphics[width=0.7\linewidth]{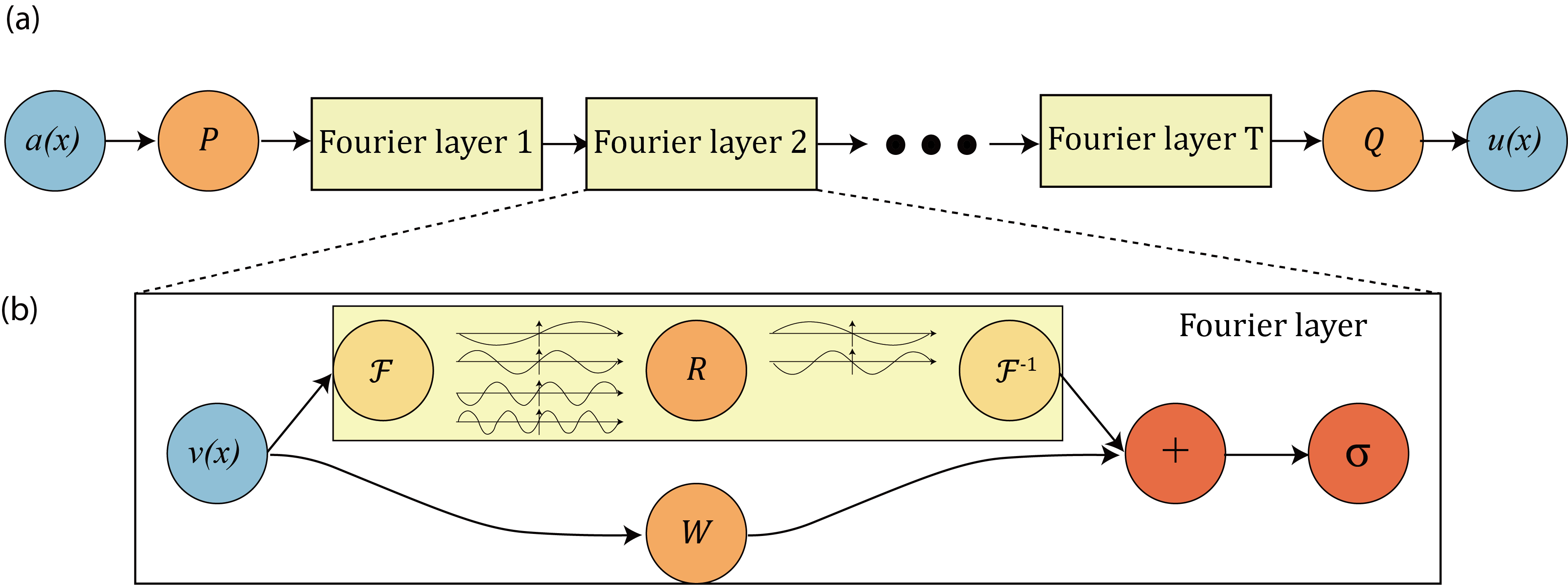}
	\caption{($a$) Full architecture of Fourier neural network. A neural network $P$ projects the input $u$ into a higher dimension; then the input goes through Fourier layers before being projected to the target dimension using a network $Q$. $u$ is the output. ($b$) is a Fourier layer that takes an input $v$. On top: applies Fourier transform $\mathcal{F}$, filters out the higher Fourier modes using a linear transform $R$, then inverse Fourier transform $\mathcal{F}^{-1}$. On bottom: applies a local linear transform $W$ \Citep{Li2021}.}%
	\label{fig:fourier}
\end{figure}

The use of neural operators has gotten its fair share of attention for seismic applications. As discussed previously in the introduction, neural operators are the second type of neural simulators, besides neural solvers (e.g., PINNs). Neural operators, mainly the Fourier neural operator (FNO) (Figure \ref{fig:fourier}) have been used successfully to solve several wave propagation and seismic applications problems \Citep{Li2021b, Wei2022, Li2023, Yang2021}. The main noteworthy difference between the two approaches is data requirements. While PINNs can incorporate labeled data in their training, it is not mandatory and can be formulated as a data-free problem; neural operators require a substantial amount of labeled data (usually simulation data) for their training. The use of neural operators is a relatively new concept, and its use-cases are still a few in the literature.

\subsubsection{Beyond basic equations: enhancing seismic inversion workflows.}%
\label{ssub:beyond_basic_equations_enhancing_wave_based_imaging_inversion}
Governing equations are mere building blocks that form the underlying physical principles of more advanced imaging techniques, such as full-waveform inversion (FWI) and wavefield reconstruction inversion (WRI). FWI is a seismic imaging technique that aims to estimate subsurface properties by iteratively matching the recorded seismic waveforms with modeled data. WRI is another advanced technique that was developed to address limitations of FWI such as computation requirements and scenarios with incomplete seismic data. Sometimes, these techniques fail to yield a unique solution (e.g., FWI as a non-linear inversion method can be trapped in local minima, especially with a bad initial model), besides the computational cost, which scales with the size and dimensions of the model. Incorporating prior geological and physical knowledge can significantly reduce the space of possible solutions by enforcing geological plausibility and fastens the convergence towards the model most representative of the subsurface. One viable option to enforce geological knowledge is a generative neural architecture called conditional variational autoencoder (CVAE), where special conditions can be enforced in the decoder part to constrain the space of possible solutions to only geologically-plausible ones \Citep{McAliley2021}.

One of the limitations of the regular FWI technique is that it only accounts for compressional waves (P-waves). An advanced form, known as  Elastic FWI uses the full elastic wave equation and incorporates both compressional and shear waves (P- and S- waves) to account for both changes in velocity and material properties. The inclusion of S-waves complicates the inversion problem and introduces a set of challenges such as the cross-talk problem, where interference occurs between the different subsurface features during the inversion process \Citep{Operto2013}. \citet{Dhara2022} tried to overcome these challenges by developing a neural architecture termed Elastic-AdjointNet which allows for encoding elastic wave physics in the neural network and could achieve inverted models free from numerical artifacts and of better accuracy than the numerical solution.  

Another notable challenge presents itself when dealing with high-frequency seismic data, which are crucial to characterize intricate and small-scale reservoir features such as thin layers or small geological anomalies. Traditional solvers encounter challenges when dealing with such data, as they necessitate increasingly finer grids due to the smaller wavelengths. This not only demands greater computing power but also introduces various numerical instabilities. Even neural networks face limitations when learning functions with high-frequency components, in a phenomenon known as \textit{spectral bias} \Citep{Rahaman2019} where the network tends to capture the lower frequency data and struggle with higher frequencies. Frequency upscaling and neuron-splitting techniques were successfully used to mitigate this problem and predict accurate solutions for high-frequency wavefields \Citep{Huang2022b}. The network becomes adept at capturing those higher-frequency features by training a shallow and narrow PINN exlusively on low frequencies and subsequently increasing its size when upscaling the learning to higher frequencies. Another successful approach to mitigate spectral bias is the Kronecker neural network (KNN), which enables the construction of wider neural architectures with minimal increase in trainable parameters. The KNN can effectively capture high-frequency features in the data, providing a valuable solution to the challenges posed by spectral bias \Citep{Jagtap2021, binWaheed2022}.

PIML models, specifically PINNs, have been a true paradigm-shift in applications of physically enhanced data-driven solutions for geophysical inversion problems. However, most of these architectures rely exclusively on the McClulloch-Pitts neural model \cite{McCulloch1943}. Although this model, which is a simplified analogy to the biological neurons, laid the groundwork for the modern neural network theory, other models provide more flexibility and possess some interesting features. For example, we can mention the Caianiello neural model \cite{Caianiello1961} which offers many advantages, such as its capability for time-varying signal processing and efficient implementation in the frequency domain using fast Fourier transform (FFT). Moreover, Caianiello neural networks provide a deterministic approach to incorporating geophysically significant models into statistical networks for inversion tasks \Citep{Fu2001}. This area of research is still lacking with very few examples \Citep{Boateng2017, Cheng2022}. 

\begin{figure}[h]
	\centering
	\includegraphics[width=0.7\linewidth]{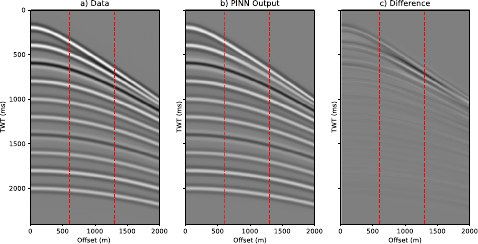}
	\caption{Seismic data interpolation using PINNs, where (a) is the original seismic data, (b) is PINN output, and (c) is their difference. Red lines denote regions with missing traces \Citep{Brandolin2022}.}
	\label{fig:interpolation}
\end{figure}

PINN-based inversion models have two main advantages over grid-based solvers \Citep{Voytan2020}, namely: 1) PINNs are mesh-free, which helps mitigate problems directly linked to grid-based numerical simulations such as numerical dispersion that is a result of discretization, and 2) during the inference stage, the entire domain is resolved (in space and time) with one forward pass of the network, which drastically decreases computational requirements. While it is unlikely these PIML models will replace traditional seismic imaging techniques soon, their implementations are advancing quickly with promising results. They can serve many purposes in the meantime, such as tools for performing quick primary analysis to accelerate decision making, seismic data interpolation (Figure \ref{fig:interpolation}) \Citep{Brandolin2022}, validating results from traditional workflows, or even providing better initial models for FWI workflows for improved convergence \Citep{Gou2023}.  

\subsection{Drilling engineering}
While a significant body of literature explores the diverse applications of neural networks in drilling engineering, the specific utilization of PIML in this field remains somewhat limited. In contrast to other domains within the oil and gas industry, such as geoscience and reservoir engineering, which heavily draw on theoretical physical knowledge, drilling engineering tends to lean more towards empirical knowledge and practical experience. As a result, the adoption and implementation of purely data-driven approaches, like vanilla neural networks, are more readily feasible in this context.

A few applications we identified when conducting our survey are the uses if PINNs to solve stick/slip problems by leveraging the physics of the drill-string system \Citep{Prasham2022, Rudat2011}, short- and long-term assessment of stuck risk for early stuck detection \Citep{Kaneko2023}, and real-time drillstring washout detection \Citep{Jan2022}. Another significant concern that is particular to offshore structures, especially drilling rigs, is vortex-induced vibrations (VIV). VIV refers to the effect of external fluid flow on the motions of bodies. This can pose certain risks on marine structures such as marine risers where the constant vibrations can induce fatigue damage. PIML models can effectively replace traditional computational fluid dynamics (CFD) simulations to model VIV systems and assess fatigue-induced damage in marine structures \Citep{Kharazmi2021, Tang2022}.

\subsection{Reservoir engineering and fluid flow}
Reservoir engineering and fluid flow analysis are critical aspects of the oil and gas industry, where accurate predictions and efficient optimization techniques are vital for maximizing production and minimizing costs. In recent years, the use of PIML approaches has shown great potential in addressing complex challenges within this area. This section aims to highlight PIML models' ability to effectively model multiphase flow behavior, simulate reservoir dynamics, optimize production strategies, and enhance decision-making processes. 

\subsubsection{Accelerating reservoir characterization and simulation}%
\label{ssub:reservoir_simulation}
Reservoir simulators are powerful and reliable tools capable of simulating even the most complex reservoirs. There are different types of reservoir simulations available, including black-oil and compositional simulations. The black-oil model, being simpler and computationally efficient, is often chosen for certain scenarios. However, when accurately modeling complex phase behavior, compositional reservoir simulation becomes indispensable. It achieves this by accounting for individual components of the hydrocarbon mixture, ensuring precise representation. In compositional simulations, flash calculations play a crucial role in determining equilibrium conditions between different phases. However, these calculations can sometimes become a bottleneck, hindering simulation efficiency. In efforts to address this issue, neural networks were augmented using governing physical laws, such as equations of state (e.g., Peng-Robinson), to perform flash calculations and predict phase behavior in the subsurface \Citep{Wu2023, Chen2023, Zhang2021, Zhang2022} or develop thermodynamically-informed neural network by incorporating thermodynamics constraint in the model \Citep{Ihunde2021}.

Flash calculations are just one example of computational bottlenecks in reservoir simulations. Another successful implementation entailed using a deep learning-based module as a proxy stress predictor in hydraulic-mechanical simulations, effectively accelerating the process \Citep{Wang2022}. This area of research still has a lot of potential, and physics-informed data-driven models incorporation as modules within numerical simulators may be the most realistic implementation that can actually have a real impact on reservoir simulation workflows in the short term, in contrast with other purely theoretical applications.

Concerning reservoir characterization, applications in the surveyed literature included porosity inversion using Caianiello neural network \Citep{Boateng2017}, self-similar and transient solutions to spontaneous imbibition by solving the governing equations \Citep{Deng2021, Abbasi2023}, and permeability prediction in various formations such as carbonate rocks and fractured coal seams \Citep{Wu2018, Tembely2021,Lv2021, Yoga2022, Grttner2023}. Permeability prediction using 3D CNN from micro-CT images represents an interesting use case that can accelerate digital core analysis workflows and rock properties characterization and address challenging formation evaluation tasks \Citep{Mellal2023}.

\subsubsection{Modeling fluid flow in porous media}
\label{ssub:fluid_flow}
\begin{figure}[htpb]
	\centering
	\includegraphics[width=0.6\linewidth]{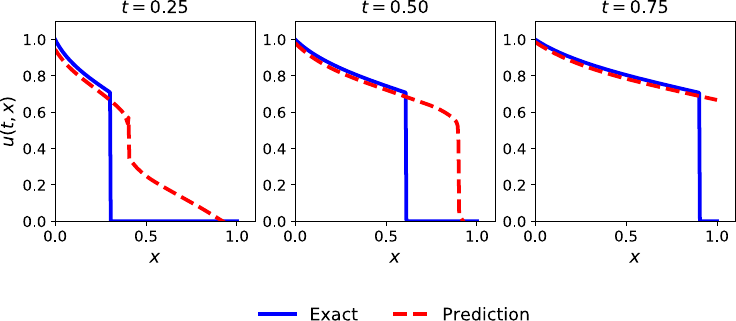}
	\caption{The predicted and exact solution to the PDE with concave flux function. The neural network prediction fails to predict the saturation front \Citep{Fuks2020}.}
	\label{fig:shock}
\end{figure}

Understanding fluid flow in porous media is paramount since it allows for the efficient exploration, production, and enhanced oil recovery techniques. The simplest case of fluid flow in porous media is single-phase flow in homogeneously permeable medium, as Darcy's law describes. Although PINNs can successfully solve such problems \Citep{ZhaoZhang2022, Sambo2021, Shen2022}, this is not really representative of most real-world scenarios. Natural world fluid flow in porous media problems entails two or more fluids (two- or multi-phase) and sometimes requires formulation as a multiphysics problem where several phenomena are involved, such as geomechanical deformations, heat transfer, and chemical reactions. Modeling two-phase fluid flow (i.e., Buckley-Leverett problem) proved non-trivial. Shocks and sharp transitions in the solution make the problem hard for the original formulation of PINNs, which can be completely unable to predict the saturation front, as shown in Figure \ref{fig:shock}  \Citep{Fuks2020, Almajid2020}. This can be mainly attributed to the discontinuity in the gradient of the hyperbolic PDE \Citep{Diab2023}. Interestingly, these problems are not unique to PINNs, but also present in numerical solvers \Citep{Diab2022}. Several strategies already presented in the literature aim to tackle this problem, such as 1) introducing attention mechanism into PINNs \Citep{Torrado2021, Diab2022}, 2) designing careful training procedures, such as decoupling the solution into non-shock components or automatically learning values for the artificial viscosity \Citep{Magzymov2021, Coutinho2023}, or 3) introducing physical-based constraints and conditions on the solution that further enforce physical laws \Citep{Gasmi2021}. Alternative architecture, such as neural operator-based simulators, which were also used to solve the multiphase flow problem, are also prone to this problem \Citep{Diab2023}. Proper uncertainty quantification can also be integrated into these models and can be crucial to assess the quality of predictions \Citep{Gasmi2023}. 

\begin{figure}[htpb]
	\centering
	\includegraphics[width=0.7\linewidth]{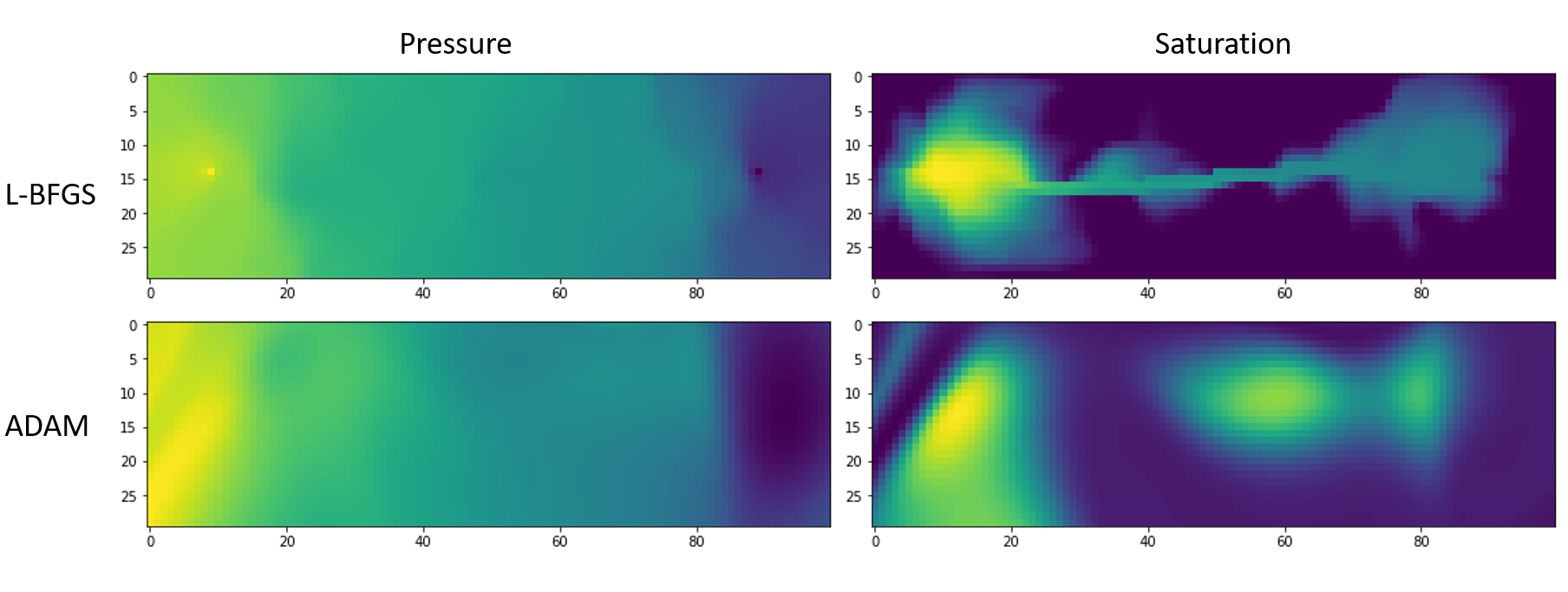}
	\caption{Pressure and saturation maps for a channelized reservoir, using both L-BFGS (top) and Adam (bottom) optimizers. L-BFGS can find a lower minimum and achieve more accurate results \Citep{Gasmi2021}.}%
	\label{fig:pressure_saturation}
\end{figure}

When trying to address multiphysics problems using PINNs, the most obvious approach would be to incorporate all the governing physics into the loss function. However, it is well known that such a complicated loss function will result in a complex loss landscape, which will make converging to a reasonable solution much harder or impossible \Citep{Haghighat2022}. \Citet{He2020} used a multiphysics PINN to solve coupled Darcy's law and the advection--dispersion law.  The authors implemented a two-phase training process: starting with a first-order optimizer (i.e., Adam), then switching to a second-order optimizer (i.e., Limited-memory Broyden--Fletcher--Goldfarb--Shanno, or L-BFGS). It is noteworthy that despite the popularity of first-order optimizers for most of the other machine learning tasks, second-order optimizers, more specifically L-BFGS, are usually the optimizers of choice for training PINNs and can effectively reach a lower error (Figure \ref{fig:pressure_saturation}) \Citep{Raissi2019,Gasmi2021, Strelow2023}. Optimizer choice and details of this two-steps training process will be discussed in more details in Section \ref{ssection:optimizer_choice}. Alternatively, \Citet{Haghighat2022} presented another example of multiphysics problems, where they coupled equations for flow and deformation of porous media, forming poromechanics problem. The authors proposed a sequential training approach based on the stress-split algorithm in poromechanics where the total loss function is split into several terms describing the different physics of the problem \Citep{Dana2021}. The sequential approach was superior to the simultaneous approach where the total loss function is used for training. Additionally, reformulating the problem in a dimensionless form resulted in a stable behavior of the optimization algorithm.

Reservoir heterogeneity For a lot of realistic fluid flow applications, we are dealing with complicated and heterogeneous reservoir conditions. Explicitly addressing these heterogeneities is crucial for proper convergence. Novel architectures such as eXtended PINN (XPINN) are specially designed to tackle this problem by performing domain decomposition and separating the whole heterogeneous domain into smaller homogeneous regions that are easier to solve \Citep{Jagtap2020b, Alhubail2022}. However, domain decomposition results in interfaces at the boundaries of each domain that need to be explicitly addressed and manually "stitched" to allow for communication across domains. Failure to model interfacial communications will affect convergence and result in inaccurate predictions.

\begin{figure}[h]
	\centering
	\includegraphics[width=.7\linewidth]{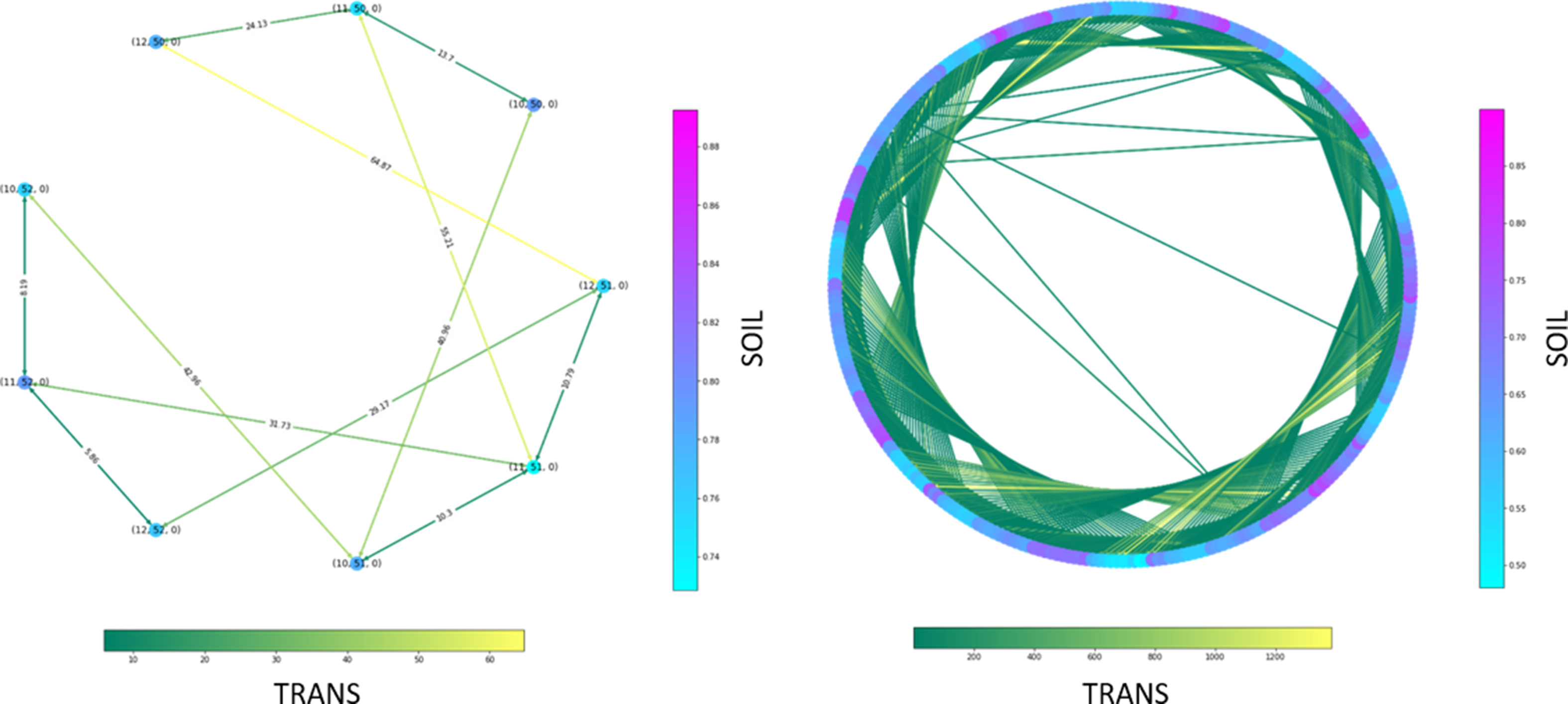}
	
	\caption{Network graph representations of a reservoir (3$\times$3$\times$1; left) and (3$\times$3$\times$10; right) simulation grids in GeoDIN architecture. Graph edges represent cell transmissibility (TRANS), and nodes represent cell oil saturation (SOIL) \Citep{Maucec2022}.}
	\label{fig:geodin}
\end{figure}

Various other architectures were used for various multi-phase fluid flow problems, such as the earlier-discussed FNO-based models \Citep{KaiZhang2022, Wen2022, Jiang2023} and interaction networks \cite{Battaglia2016}. Deep interaction networks are models that can reason about objects interaction within complex systems. They have been successfully used to simulate fluid flow by encoding reservoir grid cells as nodes in the network \Citep{Maucec2022}. An example of an interaction network for fluid flow simulation, named GeoDIN, is shown in Figure \ref{fig:geodin}.

\subsubsection{Waterflooding and enhanced oil recovery (EOR)}
The capacitance-resistance model (CRM) played a pivotal role in the oil and gas industry, being a powerful---yet simple---tool for fast production forecasting and waterflooding optimization. However, CRM can be oversimplistic to represent real-world cases for several reasons such as its oversimplified representation of the reservoir, neglect of complex flow mechanisms, and limited applicability for unconventional reservoirs. Several extensions and modifications have been developed to address these shortcomings \Citep{Holanda2018}. Although CRM mathematical formulation is too simplistic to be an accurate representation of complex reservoirs, incorporating it within the PINN had a regularizing effect and yielded promising results \Citep{Gladchenko2023, Behl2023, Nagao2023,Maniglio2021}, or by using CRM predictions as inputs to ML models \Citep{Manasipov2023}. CRM does not take into consideration the spatial distribution of wells, therefore choosing an appropriate architecture that considers the spatial layout of wells in an oil field, such as graph neural networks, and regularizing it using the CRM formulation can lead to a more intuitive representation of the subsurface dynamics, better assessment and interpretability of wells connectivity, and therefore better predictions \Citep{WendiLiu2023, Darabi2022}.

PIML use cases for simulating and optimizing other types of recovery such as EOR are still lacking, with very few examples such as optimizing steam injection operations \Citep{Sarma2017, Zhao2018} and CO\textsubscript{2} EOR \Citep{Liu2023}. It is noteworthy that the work done by \citet{Sarma2017} is very similar to the PINN formulation introduced by \citet{Raissi2019} a couple of years later. Nevertheless, their work lacked explicit mathematical development.

\subsection{Production forecasting and optimization}
Hydrocarbon production forecasting plays a crucial role as it enables operators and decision-makers to anticipate and plan for future oil and gas production levels. Accurate production forecasts are essential for optimizing reservoir management strategies, maximizing production efficiency, and ensuring the economic viability of hydrocarbon reservoirs. High-fidelity models and history matching workflows are computationally expensive, meanwhile analytical and empirical models such as decline curve analysis (DCA) are sufficient to make quick predictions that can aid with decision making, they lack the integration of the underlying subsurface structures and reservoir properties.

Combining physics and data-driven methods can be a sound approach. This is especially true for unconventional reservoirs without a complete and mature mathematical model to describe fluid flow \Citep{Shawaf2023, Ifrene2023}. Therefore, fluid flow, especially for production forecasting purposes, in unconventional reservoirs falls within the limited physical knowledge regime where data is necessary to compensate for the lack of concrete physical formulations. PINNs implementations for production forecasting scenarios with limited physics are showing promising initial results \Citep{Liu2021, Razak2022, Razak2021c, Molinari2021}, or even for enhancing established workflows such as decline curve analysis \Citep{Busby2020}, reservoir pressure management by optimizing fluid extraction rates during fluid injection operations \Citep{Harp2021}, and related tasks such as virtual flow metering (VFM) solutions \Citep{Bikmukhametov2020, Staff2020, Franklin2022}.

With the increasing reliance on unconventional resources, it is more crucial than ever to develop accurate physical models to better understand the underlying mechanisms controlling fluid flow and production from these resources. One ability of PIML that is less studied is its use to discover underlying physics. While it is unlikely that this approach will result in groundbreaking discoveries in this area, it can serve as an aiding-tool to uncover the underlying mechanisms and facilitate developing concrete and accurate physical models.

\subsection{Carbon storage}%
\label{sub:_co_2_capture_and_storage}
Carbon capture and storage (CCS) is a key strategy for reducing anthropogenic emissions of CO\textsubscript{2} and other greenhouse gases and addressing global warming. While simulating CO\textsubscript{2} storage in the underground shares similarities with traditional fluid flow problems, it possesses certain differentiating characteristics. These include the complexities of multiphase flow involving CO\textsubscript{2} and brine, the influence of capillary trapping and dissolution phenomena, geochemical reactions between CO\textsubscript{2} and reservoir rocks, and the potential for CO\textsubscript{2} plume migration and leakage risks \Citep{ZhenZhang2022b, Aminu2017}. 

PIML is proving to be a capable tool to enhance or replace reservoir simulation in the context of CCS. PINNs can be successfully employed to  model fluid flow \Citep{HonghuiDu2023, Yan2022}, predict CO\textsubscript{2} storage site response by predicting spatio-temporal evolution of the pressure and saturation \Citep{Shokouhi2021}, or identify plume migration \Citep{MingliangLiu2023}. FNO-based models were also tested and could even accurately model very complex 4D cases of CO\textsubscript{2} injection \Citep{Jiang2023} and infer sites' storage potential \Citep{Tariq2023}.

CO\textsubscript{2} flow is a complicated multiphysics problem, especially if you throw heterogeneity in the mix. Imposing closed-form residuals of highly non-linear, complex PDEs can drastically hinder convergence. \citet{Yan2022b} described a novel approach termed gradient-based deep neural network (GDNN) that they used to model CO\textsubscript{2} injection in saline aquifers. It shares similarities with traditional PINNs where automatic differentiation is used to calculate gradients of predicted state variables with respect to the inputs. However, in GDNN, the PDEs are transformed into a dictionary of elementary differential operators $\{u_t, u_x, u_{xx}, \dots\}$ (where $u$ is the state variable) that are inherently related to the physics of the fluid flow. This is the first use-case of this approach, which can be extended to address multiphysics problems in other areas since these represent a common cause of failure for PINNs.

\section{Addressing failure modes in PINNs: Strategies to overcome limitations.}
\label{section:improvement}
PINNs are prone to certain modes of failure that can hinder the model's training and convergence. Understanding their intricacies is crucial for a successful implementation. We already discussed several failure modes in various contexts and applications. Various failure modes, their reasons and remedial actions are summarized in Table \ref{table:modes-of-failure}. Next, we will elaborate on some remedial actions and discuss their advantages and implementation details.

\begingroup
\fontsize{9pt}{9pt}\selectfont
\begin{longtblr}[
  label = table:modes-of-failure,
  entry = none,
  caption = {Common failure modes in PINNs, reasons, and remedial actions.}
]{
  width = \linewidth,
  colspec = {Q[190]Q[300]Q[160]Q[230]Q[90]},
  cell{3}{1} = {r=3}{},
  cell{3}{2} = {r=3}{},
  cell{3}{3} = {r=3}{},
  cell{6}{1} = {r=2}{},
  cell{6}{2} = {r=2}{},
  cell{6}{3} = {r=2}{},
  hline{1,2,10} = {-}{0.1em},
  hline{2-3,6, 8-10} = {-}{0.03em},
  hline{4, 5, 7} = {4-5}{0.03em},
}
Failure mode                                    & Reason                                                                                               & Examples                 & Remedial actions                                                & References \\
Transient problem                                    & Learning the whole spatio-temporal domain is a hard task                             &      Highly transient problems           & Seq-to-Seq learning &          \Citep{Krishnapriyan2021}  \\
Sharp transitions / discontinuities & Discontinuity in the gradient of hyperbolic PDEs                                                                                                      & Buckley-Leverett problem & Solution decoupling                                                          &          \Citep{Magzymov2021}\\
									&                                                                                                      &                          & Attention PINN                                                          &          \Citep{Torrado2021} \Citep{Diab2022}  \\
									&                                                                                                      &                          & Physical constraints                                                          &          \Citep{Gasmi2021}  \\
Fitting high frequency functions                     & Spectral bias of neural networks                                                                     &High-frequency seismic applications                           & Kronecker NN                                 &       \Citep{Jagtap2021} \Citep{binWaheed2022}     \\
													 &                                                                                                      &                          & Frequency upscaling                                      &           \Citep{Huang2022b} \\
Multiphysics problems                      & Complicated loss landscape & Compositional reservoir modeling                          & Sequential or curriculum learning                                                          &       \Citep{Haghighat2022}     \\
Heterogeneous media                         & PINNs can fail to learn the discontinuities in the solution                                          & Fluid flow in heterogeneous media                          & Domain decomposition (XPINN)                             &            \Citep{Jagtap2020b} \\
\end{longtblr}
\endgroup

\subsection{Weighting the loss function terms}
The loss function of a PINN can have different terms, such as the PDE residual, initial and boundary conditions errors, and data error. By default, equal importance is given to the different terms; however, one can introduce weights to each term to change its importance, as is shown in Equation \ref{eq:loss_regularized}. For example, one can emphasize the PDE residual term or the data more and choose the weights $\lambda_i$ accordingly.

Besides emphasizing specific terms, regularizing the loss function in this way may be necessary to avoid the failure mode related to stiffness in the gradient flow dynamics, which leads to an imbalance in the magnitude of back-propagated gradients and hinders training \Citep{Wang2020b}. Manually choosing the loss term weights is tedious and inefficient. Alternatively, the choice of the weights $\lambda_i$ can be deferred to the training phase, where the back-propagated gradients statistics can be used to tune the weights. In doing so, the different loss terms are balanced, which was shown to result in better results \Citep{Wang2020b}.

\subsection{Optimizer choice}
\label{ssection:optimizer_choice}
We can distinguish two types of optimizers for training neural networks: 1) first-order optimizers, such as stochastic gradient descent (SGD) and Adam \Citep{Robbins1951,Kingma2017}, and 2) second-order optimizers, such as Newton and quasi-Newton family of algorithms (e.g., BFGS and L-BFGS \Citep{Liu1989}). First-order gradient-based optimizers update the model's parameters solely using first-order derivatives (gradients) of the loss function. They are widely used with proven reliability for a wide range of tasks \Citep{Schmidt2021}. On the other hand, second-order optimizers use second-order derivatives, that is the Hessian matrix (i.e., Newton methods), or its approximation (i.e., quasi-Newton methods) for computational efficiency.

\begin{figure}[htpb]
	\centering
\begin{subfigure}[b]{0.45\textwidth}
	\centering
	\includegraphics[width=\linewidth]{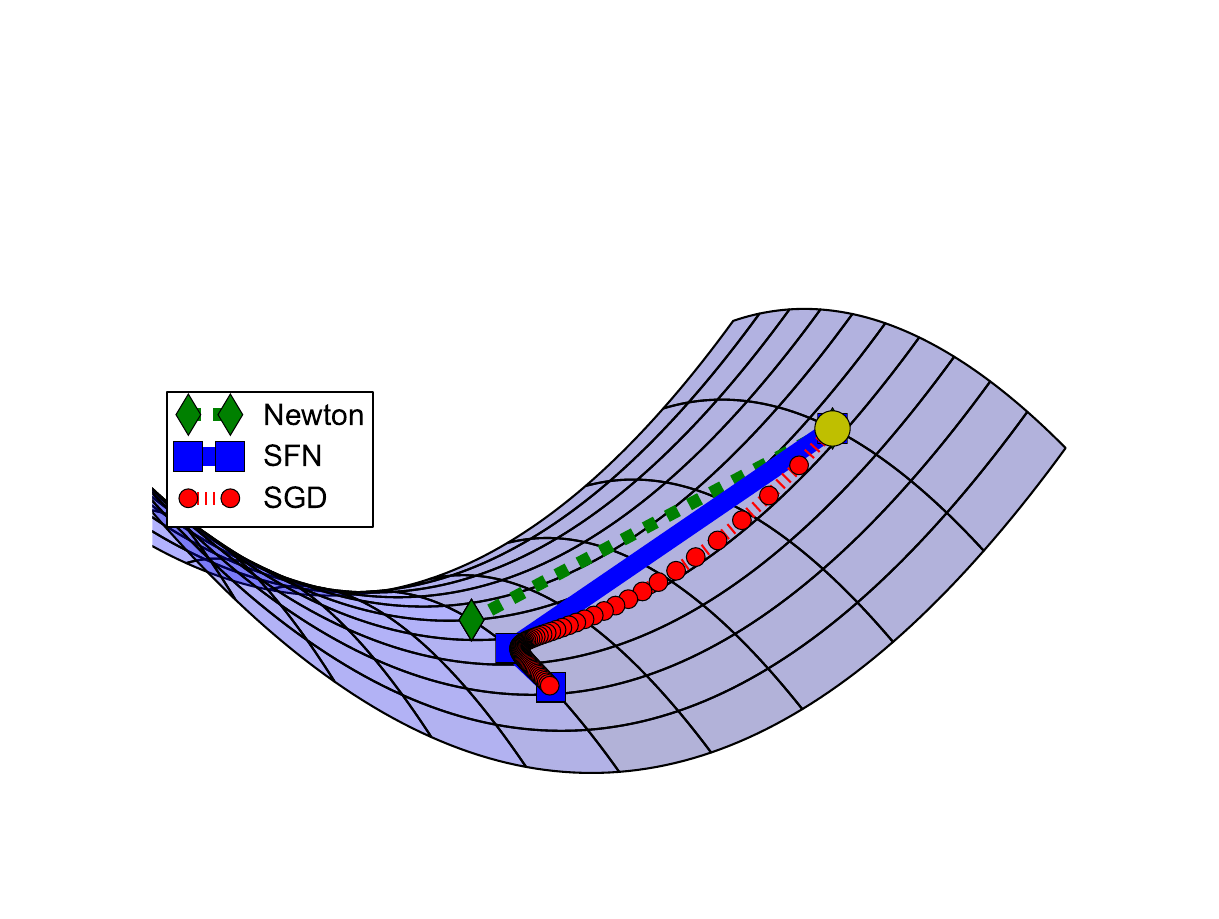}
\subcaption{}%
\end{subfigure}
\begin{subfigure}[b]{0.45\textwidth}
	\centering
		\includegraphics[width=\linewidth]{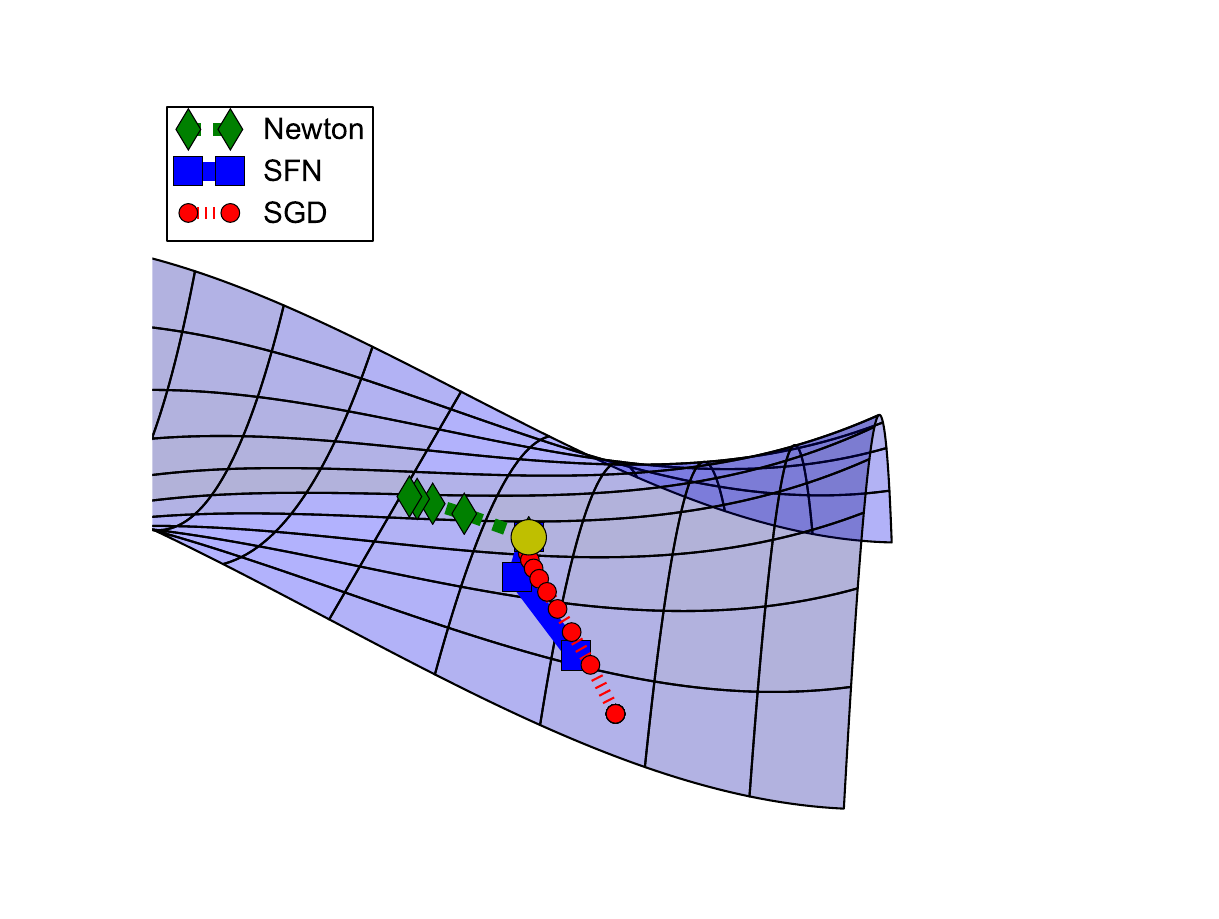}
\subcaption{}
		\end{subfigure}
		\caption{Behavior of Newton method, SFN, and SGD near (a) classical saddle structure and (b) monkey saddle structure. The yellow dot is the starting point \Citep{Dauphin2014}.}
		\label{fig:saddle_behavior}

\end{figure}

The challenge with training deep neural networks is the highly non-convex nature of the loss landscape. This poses challenges such as getting stuck in local minima and saddle points. It is becoming more apparent that local minima are less of a concern for training deep neural networks, since the deeper the network gets, the local minima are of higher quality and are more concentrated around the global minimum \Citep{Pascanu2014, Choromanska2015}. However, the proliferation of saddle points poses an actual problem for training deep networks, which are surrounded by high-error regions that can slow down training and give the impression of a local minimum \Citep{Pascanu2014}. Gradient-based optimizers with momentum (e.g., Adam) may be less prone to converge to bad local minima \Citep{Sutskever2013} and can effectively avoid saddle points \Citep{Lee2017}. On the other hand, second-order methods do not treat saddle points appropriately, and they can be attractive to these methods. It is worth mentioning that some newer variations, such as the saddle-free Newton method (SFN) can effectively escape high dimensional saddle points (Figure \ref{fig:saddle_behavior}).

Regarding PINN implementations, L-BFGS seems to be the optimizer of choice in most of the surveyed literature and frequently outperforms first-order optimizers \Citep{Raissi2019, Gasmi2021, Almajid2020, Strelow2023, Song2021, He2020}. Certain works argue for the use of a two-steps training process that stars with a first-order optimizer for the first few epochs---which are much faster and more robust against local minima and saddle points, then switch to L-BFGS for the remainder of the training process to further fine-tune and refine the solution \Citep{Markidis2021, Strelow2023, He2020}. This can be true in theory, especially for highly non-convex problems such as multiphysics problems with complex loss functions.

\subsection{Activation functions}
Nonlinear activation functions are crucial for neural architectures since they introduce nonlinearity to the model. The lack thereof will reduce the network to a simple linear transformation, insufficient to learn complex patterns. Rectified Linear Unit (ReLU), Leaky ReLU, sigmoid, and tanh are usually used. ReLU and its derivatives have gained popularity due to their properties such as computational simplicity and helping with avoiding vanishing gradients. Nevertheless, the loss function in PINNs usually involves higher-order derivatives, and ReLU is not a viable solution since its second derivative is always null. Some ReLU-like activation functions that can be suitable for training PINNs and for replacing sigmoid and tanh are swish and Gaussian error linear unit (GELU) \Citep{Hendrycks2023}. Swish has shown improved results over other functions for many tasks, including solving wave propagation problems  \Citep{Alsafwan2021, Eger2019}. Meanwhile GELU has gained popularity lately, especially for training large language models, and can be another viable, good option.

\begin{figure}[h]
    \centering
	\includegraphics[width=\linewidth]{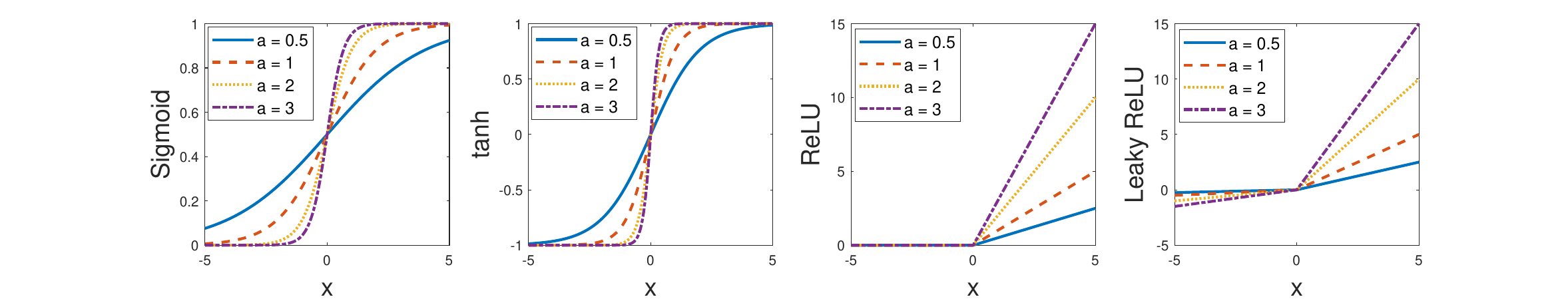}
	\caption{From left to right: Sigmoid, tanh, ReLU, Leaky-ReLU activation functions for various values of the parameter $a$ \Citep{Jagtap2020}.}
    \label{fig:adaptive}
\end{figure}

Adaptive activation functions involve introducing a learnable hyperparameter ($a$) that changes the slope of the activation function (Figure \ref{fig:adaptive}) and needs to be optimized along with the weights and biases to minimize the loss \Citep{Jagtap2020}. This is a generalization of Parametric ReLU to other types of activation functions \Citep{He2015}. Adaptive activation functions yielded excellent results in a variety of tasks, such as wave propagation problems \Citep{Jagtap2020, Jagtap2021} and fluid flow \Citep{Sambo2021}.

\subsection{Sequence-to-sequence learning}
Highly transient problems are hard to solve, and training the model on the entire space and time domains at once will probably make converging impossible. In sequence-to-sequence learning \Citep{Krishnapriyan2021}, the time domain, $T$, can be split into $n$ time steps $\{\Delta T_1, \Delta T_2, \dots \Delta T_n\}$, and the model learns to predict the solution for only $t \in \Delta T_1$, then progress to predict the solution for the next time sequence, up to $\Delta T_n$. This temporal domain decomposition makes the optimization task much easier and enables the model to converge to an accurate solution.

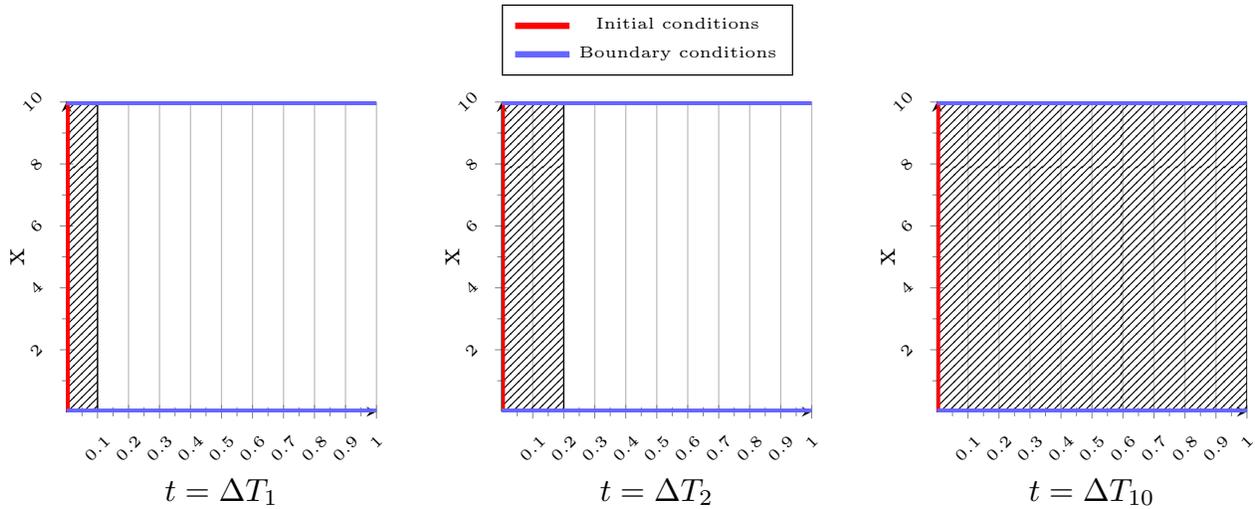
\begin{figure}[h]
    \centering
	\begin{subfigure}[b]{0.32\textwidth}
	\resizebox{\linewidth}{!}{%
		\begin{tikzpicture}
		\begin{axis}[ 
		xmin=0, xmax=1,ymin=0, 
		ymax=10, xmajorgrids,xtick = {0, .1,...,1},
		height=5cm, width=5cm,
		axis line style={latex-latex},
		axis lines=middle,
		ticklabel style={font=\tiny, rotate=45},
		minor tick num=1,
		xlabel={$t=\Delta T_{1}$},
		ylabel=x,
		    y label style={at={(axis description cs:-0.1,.5)},rotate=90,anchor=south},
		x label style={at={(axis description cs:0.5,-0.18)},anchor=north},
		scaled ticks=false] 
		\filldraw [pattern=north east lines] (rel axis cs:0,0) rectangle (.1,10);
		\addplot [color=red, line width=2pt, mark options={fill=red!80!white}] coordinates {(0,0) (0,10)};
		\addplot [color=white!40!blue, line width=2pt, mark options={fill=blue!80!white}] coordinates {(0,0) (1,0)};
		\addplot [color=white!40!blue, line width=2pt, mark options={fill=blue!80!white}] coordinates {(0,10) (1,10)};

		\end{axis}
	\end{tikzpicture} 
	}%
\end{subfigure}
\hfill
\begin{subfigure}[b]{0.32\textwidth}
	\resizebox{\linewidth}{!}{%
		\begin{tikzpicture}
		\begin{axis}[ 
		xmin=0, xmax=1,ymin=0, 
		ymax=10, xmajorgrids,xtick = {0, .1,...,1},
		height=5cm, width=5cm,
		axis line style={latex-latex},
		axis lines=middle,
		ticklabel style={font=\tiny, rotate=45},
		minor tick num=1,
		xlabel={$t=\Delta T_{2}$},
		ylabel=x,
		    y label style={at={(axis description cs:-0.1,.5)},rotate=90,anchor=south},
		x label style={at={(axis description cs:0.5,-0.18)},anchor=north},
		scaled ticks=false, legend columns=1, legend style={at={(0,1.2)},anchor=west}] 
		\filldraw [pattern=north east lines] (rel axis cs:0,0) rectangle (.2,10);
		\addplot [color=red, line width=2pt, mark options={fill=red!80!white}] coordinates {(0,0) (0,10)};
		\addlegendentry{\tiny Initial conditions}
		\addplot [color=white!40!blue, line width=2pt, mark options={fill=blue!80!white}] coordinates {(0,0) (1,0)};
		\addlegendentry{\tiny Boundary conditions}
		\addplot [color=white!40!blue, line width=2pt, mark options={fill=blue!80!white}] coordinates {(0,10) (1,10)};

		\end{axis}
	\end{tikzpicture} 
	}%
\end{subfigure}
\hfill
		\begin{subfigure}[b]{0.32\textwidth}
			\resizebox{\linewidth}{!}{%
		\begin{tikzpicture}
		\begin{axis}[ 
		xmin=0, xmax=1,ymin=0, 
		ymax=10, xmajorgrids,xtick = {0, .1,...,1},
		height=5cm, width=5cm,
		axis line style={latex-latex},
		axis lines=middle,
		ticklabel style={font=\tiny, rotate=45},
		minor tick num=1,
		xlabel={$t=\Delta T_{10}$},
		ylabel=x,
		    y label style={at={(axis description cs:-0.1,.5)},rotate=90,anchor=south},
		x label style={at={(axis description cs:0.5,-0.18)},anchor=north},
		scaled ticks=false] 
		\filldraw [pattern=north east lines] (rel axis cs:0,0) rectangle (1,10);
		\addplot [color=red, line width=2pt, mark options={fill=red!80!white}] coordinates {(0,0) (0,10)};
		\addplot [color=white!40!blue, line width=2pt, mark options={fill=blue!80!white}] coordinates {(0,0) (1,0)};
		\addplot [color=white!40!blue, line width=2pt, mark options={fill=blue!80!white}] coordinates {(0,10) (1,10)};

		\end{axis}
	\end{tikzpicture} 
		}%
		\end{subfigure}
	\caption{Sequence-to-sequence learning with the time domain split into 10 segments (adapted from \Citep{Krishnapriyan2021}).}
    \label{fig:seq2seq}
\end{figure}

\subsection{Curriculum learning}
\label{ssub:curriculum}
Some tasks may prove hard to learn for neural networks in general, and we have discussed several examples in the context of PINNs where the models can struggle to solve certain problems. \textit{Curriculum learning} is a machine learning technique that mimics the learning process of humans by training the model on easy data and progressing to more complex ones \Citep{Bengio2009}.

\citet{Krishnapriyan2021} discussed curriculum learning in the context of PIML. The problem the authors encountered is the model's inability to solve for one-dimensional convection problem characterized by a high convection coefficient. To address this, they started the training process with a significantly small value for the coefficient, which is much easier to solve, then gradually increasing it, thus increasing the problem's difficulty until the solution is reached with the required high value. This was also used in a different form by \citet{Tang2022} and was referred to as stepwise iterative training. Curriculum learning is such a simple solution that can be easily implemented and can potentially improve performance on a wide range of problems.


\section{Recap, implications, and prospects}
\label{section:conclusion}

Deep learning has revolutionized various domains and is poised to play a crucial role in the coming decades. However, the unique requirements of scientific applications often hinder the applicability of general machine learning and deep learning models. In response, a novel approach called Physics-Informed Machine Learning has emerged, fusing physical knowledge and data-driven models. This integration results in physically informed data-driven models that exhibit robustness for scientific applications and avoid limitations associated with traditional architectures such as physically insensible predictions or the inability to extrapolate beyond distribution of the training data \Citep{Torrado2021, Alkhalifah2021}.  In the context of the oil and gas industry, PIML, specifically PINNs, has proven successful in various tasks. In geophysics, it was used for modeling wave propagation and seismic inversion. In drilling engineering, it has applications for washout detection, stick/slip detection, and VIV modeling. In reservoir engineering, PIML proved to be a reliable tool for proxy modeling, single- and multi-phase flow modeling, among others. It was also used for waterflooding optimization, production forecasting, and modeling underground carbon storage.

Despite the promising results, PIML models do have known limitations. Challenges arise when solving governing equations with sharp transitions in the solution, or when tackling multiphysics problems. However, the field is witnessing the development of innovative methodologies to address these shortcomings. We have seen the use of XPINNs for heterogeneous media, Kronecker neural networks for addressing spectral bias, and the introduction of neural operator-based architecture. Existing literature often focuses on relatively simple problems that may not fully reflect real-world applications. Nonetheless, the rapid rate of advancement in this field instills optimism for the future. The most realistic obstacle might be the difficulty of implementation, since properly developing PIML-based solutions requires both domain and deep learning knowledge.

Although PIML will not replace high-fidelity grid-based numerical simulators any time soon, it provides us with powerful tools that can be developed and deployed in a fraction of the time and computational cost required by traditional simulations. We speculate that these models will serve as a tool to provide quick and rough estimates for decision-making and operations optimization. They can also help accelerate numerical simulators in several ways, such as predicting initial models to reduce the time required to converge to a solution or replace certain modules where they can provide accurate and very fast predictions compared to legacy code.

Examining Tables \ref{table:summary} and \ref{table:applications} (Appendix A), we can identify areas of potential research. Literature is abundant on geophysical applications, which can be attributed to the strong theoretical foundation of the field, while drilling engineering applications are quite limited for the lack thereof. PIML's ability to discover governing equations can be used to enhance our understanding of the physics of fluid flow and production in unconventional reservoirs. There is room for a lot of innovation in the context of underground carbon and hydrogen storage. Conversely, underground storage for carbon and hydrogen is still relatively new compared to other areas that are already well-established and well-understood. We can see how PIML can be used to simulate many scenarios for potential storage sites and identify the most suitable locations.

\begin{landscape}
\section*{Appendix A}
\begin{longtblr}[
  label = table:applications,
  entry = none,
  caption = {Summary of a few applications, approaches, and findings.}
]{
  rowhead=0,
  rowfoot=0,
	  width = \linewidth,
  colspec = {Q[30]Q[170]Q[190]Q[430]Q[145]},
  cell{2}{1} = {r=8}{},
  cell{2}{2} = {r=4}{},
  cell{2}{4} = {r=4}{},
  cell{6}{2} = {r=2}{},
  cell{10}{1} = {r=4}{},
  cell{14}{1} = {r=4}{},
  cell{18}{1} = {r=7}{},
  cell{19}{2} = {r=4}{},
  cell{19}{4} = {r=4}{},
  cell{25}{1} = {r=3}{},
  cell{25}{2} = {r=2}{},
  cell{28}{1} = {r=3}{},
  cell{28}{2} = {r=2}{},
  cell{31}{1} = {r=5}{},
  cell{32}{2} = {r=3}{},
  cell{32}{4} = {r=2}{},
  hline{10,14,18,25,28,31} = {-}{0.03em},
  hline{3-5,20-22, 33} = {3,5}{0.03em},
  hline{6,8-9,11-13,15-17,19,23, 24,27,30,32,34,35} = {2-5}{0.03em},
  hline{7,19,26,29,32} = {3-5}{0.03em},
  hline{ 2, 36} = {-}{0.1em},
  hline{1} = {2-5}{0.1em},
}
                      & Application                                              & Approach                                                      & Findings                                                                                                                                                                                 & References \\

\STAB{\rotatebox[origin=c]{90}{Geoscience}}           & Wave propagation                                         & PINN with acoustic wave eq.                              & Solve the wave equation and its derivatives by imposing        them in the loss function, or using neural operators, to obtain velocity and density fields and model     the subsurface &    \Citep{Moseley2020} \Citep{Kumar2020} \Citep{Voytan2020} \Citep{Xu2019} \Citep{YijieZhang2023}    \\
                      &                                                          & PINN with Helmholtz eq.                                  &                                                                                                                                                                                         &   \Citep{Alkhalifah2021} \Citep{Alkhalifah2020b} \Citep{Song2021} \Citep{Song2020}  \Citep{Alkhalifah2020}\\
                      &                                                          & PINN with eikonal eq.                                    &                                                                                                                                                                                         &   \Citep{binWaheed2020}  \\
					  &                                                          & Fourier neural operator                                       &                                                                                                                                                                                         &    \Citep{Wei2022} \Citep{Li2023} \Citep{Yang2021} \\
					  & High-frequency seismic data and overcoming spectral bias & Kronecker neural network                                      & Construction of larger network with minimal increase in trainable parameters allowed for overcoming spectral bias                                                             &   \Citep{binWaheed2022}  \\
					  &                                                          & Frequency upscaling and neuron splitting                                           & Using lower frequencies for training followed by higher ones, combined with neuron splitting mitigates spectral bias.                                                       &    \Citep{Huang2022b} \\
					  & Data interpolation                                       & PWD-PINN                                                 & PINN trained to solve the local plane wave differential equations and the local slope could interpolate seismic data&    \Citep{Brandolin2022} \\
					  & Seismic source location                                  & PINN with Helmholtz eq.                                                 & Locate subsurface seismic sources with high accuracy                                                                                                                                                                                 &    \Citep{Grubas2021} \Citep{Huang2022} \\
\STAB{\rotatebox[origin=c]{90}{Drilling}}              & Stick/slip detection                                               & Drillstring physical-model integration& Improved prediction of stick/slip severity compared to traditional models                                                                                                               &   \Citep{Prasham2022}  \\
																		& Early stuck detection                                    & Feature engineering & Reliably predict short- and long-term stuck risk                                                                                                                                        &    \Citep{Kaneko2023} \\
																		& Washout detection                                        & Multitasking PINN                                                               & Detect drillstring washout from real-time field measurements                                                                                                                                                                                         &    \Citep{Jan2022} \\
					  & VIV modeling                                             & Digital twins and PINNs                                                               & Replace CFD simulations to model VIV and assess fatigue &     \Citep{Kharazmi2021} \Citep{Tang2022}\\
\STAB{\rotatebox[origin=c]{90}{\makecell[b]{Reservoir Charac.\\\& Simulation}}} & Flash calculations                                       & PINNs and TINNs                                                               & Accelerate phase equilibrium calculations for compositional reservoir modeling                                                                                                                                                                                         &    \Citep{Wu2023} \Citep{Chen2023} \Citep{Ihunde2021} \Citep{Zhang2021} \Citep{Zhang2022} \\
                      & Proxy modeling                                           & GeoDIN                                                               & Interaction networks can effectively replace numerical models                                                                                                                                                                                         &   \Citep{Maucec2022}  \\
                      & Permeability prediction                                           & 3D CNN                                                               & Accurate permeability estimation from micro-CT images                                                                                                                                                                                         &   \Citep{Tembely2021}  \\
					  & Spontaneous imbibition & PINNs                                                               & Accurate solution for self-similar and transient spontaneous imbibition                                                                                                                                                                                         &   \Citep{Abbasi2023}  \Citep{Deng2021}\\

\STAB{\rotatebox[origin=c]{90}{Fluid flow}}           & Single-phase fluid flow                                  &PINNs with Darcy's law                                                               & Accurate solution for single-phase fluid flow &    \Citep{ZhaoZhang2022} \Citep{Sambo2021} \Citep{Shen2022} \\
					   & Multi-phase fluid flow                                   & Attention PINN                                                & Introducing attention mechanisms, solution decoupling, physical constraints and adaptive AV allowed to capture sharp discontinuities in the solution that can't be solved with plain PINNs \Citep{Fuks2020}                                                                                                                                                                                         &    \Citep{Torrado2021} \Citep{Diab2022} \\
					  &                                                          & Solution decoupling                                           &                                                                                                                                                                                         &   \Citep{Magzymov2021}  \\
					  &                                                          & Physical constraints                              &                                                                                                                                                                                         &   \Citep{Gasmi2021}  \\
					  &                                                          & Adaptive AV                                &                                                                                                                                                                                         &   \Citep{Coutinho2023}  \\
					  					  & Heterogeneous media                                      & eXtended PINN                                                 & Accurate modeling for heterogeneous reservoirs                                                                                                                                                                                         &    \Citep{Alhubail2022} \\
					  & Multiphysics simulation                                  & PINN + coupled-physics loss                                                               & Sequential training may be needed for convergence                                                                                                                                                                                         &  \Citep{Haghighat2022} \Citep{He2020}   \\
\STAB{\rotatebox[origin=c]{90}{\makecell[b]{Waterflooding\\\& EOR}}}& Waterflooding optimization                               & PINN with CRM formulation                                     & Regularizing PINNs with CRM formulation can yield better results beyond traditional CRM                                                                                                                                                                                         &    \Citep{Gladchenko2023} \Citep{Behl2023} \Citep{Nagao2023} \Citep{Maniglio2021} \\
														&                                                          & Graph PINN                                                    & Graphs are a better representation for waterflooding settings, especially when coupled with physical constraints                                                                                                                                                                                         &    \Citep{WendiLiu2023} \Citep{Darabi2022} \\
					  & Steam injection                                          & Data physics (see discussion for details)                                                                &  \textit{Data physics} approach is comparable to numerical simulators in accuracy but at a fraction of the computational time                                                                                                                                                                                        &    \Citep{Sarma2017} \Citep{Zhao2018} \\
{\STAB{\rotatebox[origin=c]{90}{Production}}}           & Forecasting                                              & PINN with flow physics                                                               & Incorporating (limited) physics of flow in PINN                                                                                                                           &     \Citep{Razak2022} \Citep{Razak2021c}\Citep{Molinari2021} \\
                      &                                                          & PINN with DCA                                                 & Regulating PINNs with DCA formulation can yield better results                                                                                                                                                                                         &     \Citep{Busby2020}\\
					  & Virtual flow metering                                    & Feature engineering and PINNs                                                               & PIML-based VFM can surpass purely data-driven solutions and replace expensive hardware                                                                                                                                                                                         &     \Citep{Staff2020} \Citep{Franklin2022} \\
\STAB{\rotatebox[origin=c]{90}{\makecell[b]{CO\textsubscript{2} Storage}}}                & Storage site response                                  & PINN                                                               &  Predict spatio-temporal pressure and saturation of CO\textsubscript{2} in saline aquifers                                                                                                                                                                                       &    \Citep{Tariq2023} \Citep{Shokouhi2021}\\
																	  & Injection and flow                                       & PINN                                                               & Both PINNs and FNO-based architecture could accurately model 4D subsurface CO\textsubscript{2} flow                                                                                                                                                                                  &  \Citep{HonghuiDu2023} \Citep{Yan2022}   \\ 
																	   &                                                          & Fourier-MIONet&                                                                                                                                                                                         &   \Citep{Jiang2023}  \\
					  &                                                          & GDNN                            &  Use of elementary differential operators instead of governing equations to solve problems with ill-posed physics                                                                                                                                                                                        &    \Citep{Yan2022b} \\
					  & Plume migration                                          & Differentiable PINN                                                               & Joint inversion of geophysical data for estimation of reservoir properties and CO\textsubscript{2} storage monitoring                                                                                                                                                                                        &     \Citep{MingliangLiu2023}\\
                      &                                                          &                                                               &                                                                                                                                                                                         &     \\
                      &                                                          &                                                               &                                                                                                                                                                                         &     \\
\end{longtblr}
\end{landscape}

\bibliography{misc/bibliography} 

\begin{thebibliography}{163}
\providecommand{\natexlab}[1]{#1}
\providecommand{\url}[1]{\texttt{#1}}
\expandafter\ifx\csname urlstyle\endcsname\relax
  \providecommand{\doi}[1]{doi: #1}\else
  \providecommand{\doi}{doi: \begingroup \urlstyle{rm}\Url}\fi

\bibitem[LeCun et~al.(2015)LeCun, Bengio, and Hinton]{LeCun2015}
Yann LeCun, Yoshua Bengio, and Geoffrey Hinton.
\newblock Deep learning.
\newblock \emph{Nature}, 521\penalty0 (7553):\penalty0 436--444, May 2015.
\newblock \doi{10.1038/nature14539}.
\newblock URL \url{https://doi.org/10.1038/nature14539}.

\bibitem[Brown et~al.(2020)Brown, Mann, Ryder, Subbiah, Kaplan, Dhariwal,
  Neelakantan, Shyam, Sastry, Askell, Agarwal, Herbert-Voss, Krueger, Henighan,
  Child, Ramesh, Ziegler, Wu, Winter, Hesse, Chen, Sigler, Litwin, Gray, Chess,
  Clark, Berner, McCandlish, Radford, Sutskever, and Amodei]{Brown2020}
Tom Brown, Benjamin Mann, Nick Ryder, Melanie Subbiah, Jared~D Kaplan, Prafulla
  Dhariwal, Arvind Neelakantan, Pranav Shyam, Girish Sastry, Amanda Askell,
  Sandhini Agarwal, Ariel Herbert-Voss, Gretchen Krueger, Tom Henighan, Rewon
  Child, Aditya Ramesh, Daniel Ziegler, Jeffrey Wu, Clemens Winter, Chris
  Hesse, Mark Chen, Eric Sigler, Mateusz Litwin, Scott Gray, Benjamin Chess,
  Jack Clark, Christopher Berner, Sam McCandlish, Alec Radford, Ilya Sutskever,
  and Dario Amodei.
\newblock Language models are few-shot learners.
\newblock In H.~Larochelle, M.~Ranzato, R.~Hadsell, M.F. Balcan, and H.~Lin,
  editors, \emph{Advances in Neural Information Processing Systems}, volume~33,
  pages 1877--1901. Curran Associates, Inc., 2020.
\newblock URL
  \url{https://proceedings.neurips.cc/paper_files/paper/2020/file/1457c0d6bfcb4967418bfb8ac142f64a-Paper.pdf}.

\bibitem[Huang et~al.(2020)Huang, Lin, Tong, Hu, Zhang, Iwamoto, Han, Chen, and
  Wu]{Huang2020}
Huimin Huang, Lanfen Lin, Ruofeng Tong, Hongjie Hu, Qiaowei Zhang, Yutaro
  Iwamoto, Xianhua Han, Yen-Wei Chen, and Jian Wu.
\newblock Unet 3+: A full-scale connected unet for medical image segmentation.
\newblock In \emph{ICASSP 2020 - 2020 IEEE International Conference on
  Acoustics, Speech and Signal Processing (ICASSP)}, pages 1055--1059, 2020.
\newblock \doi{10.1109/ICASSP40776.2020.9053405}.

\bibitem[Li et~al.(2018)Li, Chen, Qi, Dou, Fu, and Heng]{Li2018}
Xiaomeng Li, Hao Chen, Xiaojuan Qi, Qi~Dou, Chi-Wing Fu, and Pheng-Ann Heng.
\newblock H-denseunet: Hybrid densely connected unet for liver and tumor
  segmentation from ct volumes.
\newblock \emph{IEEE Transactions on Medical Imaging}, 37\penalty0
  (12):\penalty0 2663--2674, 2018.
\newblock \doi{10.1109/TMI.2018.2845918}.

\bibitem[Selvaraju et~al.(2019)Selvaraju, Cogswell, Das, Vedantam, Parikh, and
  Batra]{Selvaraju2019}
Ramprasaath~R. Selvaraju, Michael Cogswell, Abhishek Das, Ramakrishna Vedantam,
  Devi Parikh, and Dhruv Batra.
\newblock Grad-{CAM}: Visual explanations from deep networks via gradient-based
  localization.
\newblock \emph{International Journal of Computer Vision}, 128\penalty0
  (2):\penalty0 336--359, oct 2019.
\newblock \doi{10.1007/s11263-019-01228-7}.
\newblock URL \url{https://doi.org/10.1007%2Fs11263-019-01228-7}.

\bibitem[Hassanin et~al.(2022)Hassanin, Anwar, Radwan, Khan, and
  Mian]{Hassanin2022}
Mohammed Hassanin, Saeed Anwar, Ibrahim Radwan, Fahad~S Khan, and Ajmal Mian.
\newblock Visual attention methods in deep learning: An in-depth survey, 2022.

\bibitem[Battaglia et~al.(2018)Battaglia, Hamrick, Bapst, Sanchez-Gonzalez,
  Zambaldi, Malinowski, Tacchetti, Raposo, Santoro, Faulkner, Gulcehre, Song,
  Ballard, Gilmer, Dahl, Vaswani, Allen, Nash, Langston, Dyer, Heess, Wierstra,
  Kohli, Botvinick, Vinyals, Li, and Pascanu]{Battaglia2018}
Peter~W. Battaglia, Jessica~B. Hamrick, Victor Bapst, Alvaro Sanchez-Gonzalez,
  Vinicius Zambaldi, Mateusz Malinowski, Andrea Tacchetti, David Raposo, Adam
  Santoro, Ryan Faulkner, Caglar Gulcehre, Francis Song, Andrew Ballard, Justin
  Gilmer, George Dahl, Ashish Vaswani, Kelsey Allen, Charles Nash, Victoria
  Langston, Chris Dyer, Nicolas Heess, Daan Wierstra, Pushmeet Kohli, Matt
  Botvinick, Oriol Vinyals, Yujia Li, and Razvan Pascanu.
\newblock Relational inductive biases, deep learning, and graph networks, 2018.

\bibitem[Karniadakis et~al.(2021)Karniadakis, Kevrekidis, Lu, Perdikaris, Wang,
  and Yang]{Karniadakis2021}
George~Em Karniadakis, Ioannis~G. Kevrekidis, Lu~Lu, Paris Perdikaris, Sifan
  Wang, and Liu Yang.
\newblock Physics-informed machine learning.
\newblock \emph{Nature Reviews Physics}, 3:\penalty0 422--440, 6 2021.
\newblock ISSN 25225820.
\newblock \doi{10.1038/s42254-021-00314-5}.

\bibitem[Hao et~al.(2023)Hao, Liu, Zhang, Ying, Feng, Su, and Zhu]{Hao2023}
Zhongkai Hao, Songming Liu, Yichi Zhang, Chengyang Ying, Yao Feng, Hang Su, and
  Jun Zhu.
\newblock Physics-informed machine learning: A survey on problems, methods and
  applications, 2023.

\bibitem[Hornik et~al.(1989)Hornik, Stinchcombe, and White]{Hornik1989}
Kurt Hornik, Maxwell Stinchcombe, and Halbert White.
\newblock Multilayer feedforward networks are universal approximators.
\newblock \emph{Neural Networks}, 2\penalty0 (5):\penalty0 359--366, January
  1989.
\newblock \doi{10.1016/0893-6080(89)90020-8}.
\newblock URL \url{https://doi.org/10.1016/0893-6080(89)90020-8}.

\bibitem[Sutskever et~al.(2013)Sutskever, Martens, Dahl, and
  Hinton]{Sutskever2013}
Ilya Sutskever, James Martens, George Dahl, and Geoffrey Hinton.
\newblock On the importance of initialization and momentum in deep learning.
\newblock In \emph{Proceedings of the 30th International Conference on
  International Conference on Machine Learning - Volume 28}, ICML'13, page
  III–1139–III–1147. JMLR.org, 2013.

\bibitem[Willard et~al.(2020)Willard, Jia, Steinbach, Kumar, and
  Xu]{Willard2020}
Jared Willard, Xiaowei Jia, Michael Steinbach, Vipin Kumar, and Shaoming Xu.
\newblock Integrating physics-based modeling with machine learning: A survey.
\newblock 1:\penalty0 34, 2020.
\newblock \doi{10.1145/1122445.1122456}.
\newblock URL \url{https://doi.org/10.1145/1122445.1122456}.

\bibitem[Zhuang et~al.(2020)Zhuang, Qi, Duan, Xi, Zhu, Zhu, Xiong, and
  He]{Zhuang2020}
Fuzhen Zhuang, Zhiyuan Qi, Keyu Duan, Dongbo Xi, Yongchun Zhu, Hengshu Zhu, Hui
  Xiong, and Qing He.
\newblock A comprehensive survey on transfer learning, 2020.

\bibitem[Yi et~al.(2022)Yi, Zhou, Habermann, Shimada, Golyanik, Theobalt, and
  Xu]{Yi2022}
Xinyu Yi, Yuxiao Zhou, Marc Habermann, Soshi Shimada, Vladislav Golyanik,
  Christian Theobalt, and Feng Xu.
\newblock Physical inertial poser (pip): Physics-aware real-time human motion
  tracking from sparse inertial sensors, 2022.

\bibitem[Sun et~al.(2020)Sun, Niu, Innanen, Li, and Trad]{Sun2020}
Jian Sun, Zhan Niu, Kristopher~A. Innanen, Junxiao Li, and Daniel~O. Trad.
\newblock A theory-guided deep-learning formulation and optimization of seismic
  waveform inversion.
\newblock \emph{{GEOPHYSICS}}, 85\penalty0 (2):\penalty0 R87--R99, March 2020.
\newblock \doi{10.1190/geo2019-0138.1}.
\newblock URL \url{https://doi.org/10.1190/geo2019-0138.1}.

\bibitem[Daw et~al.(2019)Daw, Thomas, Carey, Read, Appling, and
  Karpatne]{Daw2019}
Arka Daw, R.~Quinn Thomas, Cayelan~C. Carey, Jordan~S. Read, Alison~P. Appling,
  and Anuj Karpatne.
\newblock Physics-guided architecture (pga) of neural networks for quantifying
  uncertainty in lake temperature modeling, 2019.

\bibitem[Ling et~al.(2016)Ling, Kurzawski, and Templeton]{Ling2016}
Julia Ling, Andrew Kurzawski, and Jeremy Templeton.
\newblock Reynolds averaged turbulence modelling using deep neural networks
  with embedded invariance.
\newblock \emph{Journal of Fluid Mechanics}, 807:\penalty0 155–166, 2016.
\newblock \doi{10.1017/jfm.2016.615}.

\bibitem[Raissi et~al.(2019)Raissi, Perdikaris, and Karniadakis]{Raissi2019}
M.~Raissi, P.~Perdikaris, and G.~E. Karniadakis.
\newblock Physics-informed neural networks: A deep learning framework for
  solving forward and inverse problems involving nonlinear partial differential
  equations.
\newblock \emph{Journal of Computational Physics}, 378:\penalty0 686--707, 2
  2019.
\newblock ISSN 0021-9991.
\newblock \doi{10.1016/J.JCP.2018.10.045}.

\bibitem[Liu and Pyrcz(2023)]{WendiLiu2023}
Wendi Liu and Michael~J. Pyrcz.
\newblock Physics-informed graph neural network for spatial-temporal production
  forecasting.
\newblock \emph{Geoenergy Science and Engineering}, 223:\penalty0 211486, 2023.
\newblock ISSN 2949-8910.
\newblock \doi{https://doi.org/10.1016/j.geoen.2023.211486}.
\newblock URL
  \url{https://www.sciencedirect.com/science/article/pii/S2949891023000726}.

\bibitem[Alkinani et~al.(2019)Alkinani, Al-Hameedi, Dunn-Norman, Flori, Alsaba,
  and Amer]{Alkinani2019}
Husam~H. Alkinani, Abo Taleb~T. Al-Hameedi, Shari Dunn-Norman, Ralph~E. Flori,
  Mortadha~T. Alsaba, and Ahmed~S. Amer.
\newblock Applications of artificial neural networks in the petroleum industry:
  A review.
\newblock \emph{SPE Middle East Oil and Gas Show and Conference, MEOS,
  Proceedings}, 2019-March, 3 2019.
\newblock \doi{10.2118/195072-MS}.
\newblock URL \url{/SPEMEOS/proceedings-abstract/19MEOS/3-19MEOS/218609}.

\bibitem[Hansen et~al.(1993)Hansen, Aps, and T]{Hansen1993}
Kim~Vejlby Hansen, Danneskiold-Samsoe Aps, and Denmark~C T.
\newblock Neural networks for primary reflection identification, 9 1993.
\newblock URL \url{/SEGAM/proceedings-abstract/SEG93/All-SEG93/83230}.

\bibitem[Karrenbach et~al.(2000)Karrenbach, Essenreiter, and
  Treitel]{Karrenbach2000}
M.~Karrenbach, R.~Essenreiter, and S.~Treitel.
\newblock Multiple attenuation with attribute-based neural networks, 8 2000.
\newblock URL \url{/SEGAM/proceedings-abstract/SEG00/All-SEG00/88094}.

\bibitem[Huang et~al.(2006)Huang, Pissarenko, Chen, Lai, and Don]{Huang2006}
Kou-Yuan Huang, Jiun-De Pissarenko, Kai-Ju Chen, Hung-Lin Lai, and An-Jin Don.
\newblock Neural network for parameters determination and seismic pattern
  detection, 10 2006.
\newblock URL \url{/SEGAM/proceedings-abstract/SEG06/All-SEG06/93564}.

\bibitem[Verma et~al.(2012)Verma, Roy, Perez, Marfurt, and
  of~Oklahoma]{Verma2012}
Sumit Verma, Atish Roy, Roderick Perez, Kurt~J Marfurt, and The~U of~Oklahoma.
\newblock Mapping high frackability and high toc zones in the barnett shale:
  Supervised probabilistic neural network vs. unsupervised multi-attribute
  kohonen som, 11 2012.
\newblock URL \url{/SEGAM/proceedings-abstract/SEG12/All-SEG12/98711}.

\bibitem[Ross(2017)]{Ross2017}
Christopher Ross.
\newblock Improving resolution and clarity with neural networks, 9 2017.
\newblock URL \url{/SEGAM/proceedings-abstract/SEG17/All-SEG17/102728}.

\bibitem[Arehart(1990)]{Arehart1990}
R.~A. Arehart.
\newblock Drill-bit diagnosis with neural networks.
\newblock \emph{SPE Computer Applications}, 2:\penalty0 24--28, 7 1990.
\newblock ISSN 1064-9778.
\newblock \doi{10.2118/19558-PA}.
\newblock URL
  \url{https://onepetro.org/CA/article/2/04/24/168581/Drill-Bit-Diagnosis-With-Neural-Networks}.

\bibitem[Dashevskiy et~al.(1999)Dashevskiy, Dubinsky, and
  Macpherson]{Dashevskiy1999}
D.~Dashevskiy, V.~Dubinsky, and J.~D. Macpherson.
\newblock Application of neural networks for predictive control in drilling
  dynamics.
\newblock 10 1999.
\newblock \doi{10.2118/56442-MS}.
\newblock URL \url{/SPEATCE/proceedings-abstract/99ATCE/All-99ATCE/60178}.

\bibitem[Fruhwirth et~al.(2006)Fruhwirth, Thonhauser, and
  Mathis]{Fruhwirth2006}
R.~K. Fruhwirth, G.~Thonhauser, and W.~Mathis.
\newblock Hybrid simulation using neural networks to predict drilling
  hydraulics in real time.
\newblock 9 2006.
\newblock \doi{10.2118/103217-MS}.
\newblock URL \url{/SPEATCE/proceedings-abstract/06ATCE/All-06ATCE/140219}.

\bibitem[Moran et~al.(2010)Moran, Ibrahim, Purwanto, and Osmond]{Moran2010}
David Moran, Hani Ibrahim, Arifin Purwanto, and Jerry Osmond.
\newblock Sophisticated rop prediction technologies based on neural network
  delivers accurate drill time results.
\newblock \emph{Society of Petroleum Engineers - IADC/SPE Asia Pacific Drilling
  Technology Conference 2010}, pages 100--108, 11 2010.
\newblock \doi{10.2118/132010-MS}.
\newblock URL \url{/SPEAPDT/proceedings-abstract/10APDT/All-10APDT/100435}.

\bibitem[Elkatatny et~al.(2016)Elkatatny, Tariq, and Mahmoud]{Elkatatny2016}
Salaheldin Elkatatny, Zeeshan Tariq, and Mohamed Mahmoud.
\newblock Real time prediction of drilling fluid rheological properties using
  artificial neural networks visible mathematical model (white box).
\newblock \emph{Journal of Petroleum Science and Engineering}, 146:\penalty0
  1202--1210, 10 2016.
\newblock ISSN 0920-4105.
\newblock \doi{10.1016/J.PETROL.2016.08.021}.

\bibitem[Thomas et~al.(1995)Thomas, La, Golder, and Redmond]{Thomas1995}
Andrew~L Thomas, Paul~R La, Pointe Golder, and Wash Redmond.
\newblock Conductive fracture identification using neural networks, 6 1995.
\newblock URL \url{/ARMAUSRMS/proceedings-abstract/ARMA95/All-ARMA95/130727}.

\bibitem[Shelley(1999)]{Shelley1999}
Robert~F. Shelley.
\newblock Artificial neural networks identify restimulation candidates in the
  red oak field.
\newblock 3 1999.
\newblock \doi{10.2118/52190-MS}.
\newblock URL \url{/SPEOKOG/proceedings-abstract/99MCOS/All-99MCOS/60556}.

\bibitem[Ghahfarokhi et~al.(2018)Ghahfarokhi, Carr, Bhattacharya, Elliott,
  Shahkarami, and Martin]{Ghahfarokhi2018}
Payam~Kavousi Ghahfarokhi, Timothy Carr, Shuvajit Bhattacharya, Justin Elliott,
  Alireza Shahkarami, and Keithan Martin.
\newblock A fiber-optic assisted multilayer perceptron reservoir production
  modeling: A machine learning approach in prediction of gas production from
  the marcellus shale.
\newblock \emph{SPE/AAPG/SEG Unconventional Resources Technology Conference
  2018, URTC 2018}, pages 23--25, 7 2018.
\newblock \doi{10.15530/URTEC-2018-2902641}.
\newblock URL \url{/URTECONF/proceedings-abstract/18URTC/3-18URTC/157001}.

\bibitem[Pankaj(2018)]{Pankaj2018}
Piyush Pankaj.
\newblock Characterizing well spacing, well stacking, and well completion
  optimization in the permian basin: An improved and efficient workflow using
  cloud-based computing.
\newblock \emph{SPE/AAPG/SEG Unconventional Resources Technology Conference
  2018, URTC 2018}, pages 23--25, 7 2018.
\newblock \doi{10.15530/URTEC-2018-2876482}.
\newblock URL \url{/URTECONF/proceedings-abstract/18URTC/2-18URTC/157249}.

\bibitem[AL-Dogail et~al.(2018)AL-Dogail, Baarimah, and Basfar]{Dogail2018}
Ala~S. AL-Dogail, Salem~O. Baarimah, and Salem~A. Basfar.
\newblock Prediction of inflow performance relationship of a gas field using
  artificial intelligence techniques.
\newblock \emph{Society of Petroleum Engineers - SPE Kingdom of Saudi Arabia
  Annual Technical Symposium and Exhibition 2018, SATS 2018}, pages 23--26, 4
  2018.
\newblock \doi{10.2118/192273-MS}.
\newblock URL \url{/SPESATS/proceedings-abstract/18SATS/All-18SATS/215613}.

\bibitem[An and Moon(1993)]{An1993}
P.~An and W.~M. Moon.
\newblock Reservoir characterization using feedforward neural networks.
\newblock \emph{SEG Technical Program Expanded Abstracts}, pages 258--262,
  1993.
\newblock \doi{10.1190/1.1822454}.
\newblock URL \url{https://library.seg.org/doi/10.1190/1.1822454}.

\bibitem[Ayoub et~al.(2007)Ayoub, Raja, and Al-Marhoun]{Ayoub2007}
M.~A. Ayoub, D.~M. Raja, and M.~A. Al-Marhoun.
\newblock Evaluation of below bubble point viscosity correlations \&
  construction of a new neural network model.
\newblock \emph{SPE - Asia Pacific Oil and Gas Conference}, 1:\penalty0
  196--205, 10 2007.
\newblock \doi{10.2118/108439-MS}.
\newblock URL \url{/SPEAPOG/proceedings-abstract/07APOGCE/All-07APOGCE/142503}.

\bibitem[Elshafei and Hamada(2009)]{Elshafei2009}
M.~Elshafei and G.~M. Hamada.
\newblock Neural network identification of hydrocarbon potential of shaly sand
  reservoirs.
\newblock \emph{http://dx.doi.org/10.1080/10916460701699868}, 27:\penalty0
  72--82, 1 2009.
\newblock ISSN 10916466.
\newblock \doi{10.1080/10916460701699868}.
\newblock URL
  \url{https://www.tandfonline.com/doi/abs/10.1080/10916460701699868}.

\bibitem[Ma and Gomez(2015)]{Ma2015}
Y.~Zee Ma and Ernest Gomez.
\newblock Uses and abuses in applying neural networks for predictions in
  hydrocarbon resource evaluation.
\newblock \emph{Journal of Petroleum Science and Engineering}, 133:\penalty0
  66--75, 9 2015.
\newblock ISSN 0920-4105.
\newblock \doi{10.1016/J.PETROL.2015.05.006}.

\bibitem[Song and Alkhalifah(2020)]{Song2020}
Chao Song and Tariq Alkhalifah.
\newblock Wavefield reconstruction inversion via machine learned functions.
\newblock volume 2020-October, pages 1710--1714. Society of Exploration
  Geophysicists, 2020.
\newblock \doi{10.1190/segam2020-3427351.1}.

\bibitem[Alkhalifah et~al.(2020)Alkhalifah, Song, and Waheed]{Alkhalifah2020}
Tariq Alkhalifah, Chao Song, and Umair~Bin Waheed.
\newblock Machine learned green's functions that approximately satisfy the wave
  equation.
\newblock volume 2020-October, pages 2638--2642. Society of Exploration
  Geophysicists, 2020.
\newblock \doi{10.1190/segam2020-3421468.1}.

\bibitem[Waheed et~al.(2020)Waheed, Haghighat, and Alkhalifah]{binWaheed2020}
Umair~Bin Waheed, Ehsan Haghighat, and Tariq Alkhalifah.
\newblock Anisotropic eikonal solution using physics-informed neural networks.
\newblock \emph{SEG Technical Program Expanded Abstracts},
  2020-October:\penalty0 1566--1570, 2020.
\newblock ISSN 19494645.
\newblock \doi{10.1190/SEGAM2020-3423159.1}.
\newblock URL \url{https://library.seg.org/doi/10.1190/segam2020-3423159.1}.

\bibitem[McAliley and Li(2021)]{McAliley2021}
W.~Anderson McAliley and Yaoguo Li.
\newblock Machine learning inversion of geophysical data by a conditional
  variational autoencoder.
\newblock volume 2021-September, pages 1460--1464. Society of Exploration
  Geophysicists, 2021.
\newblock \doi{10.1190/segam2021-3594761.1}.

\bibitem[Waheed(2022)]{binWaheed2022}
Umair~Bin Waheed.
\newblock \emph{Kronecker neural networks for the win: Overcoming spectral bias
  for PINN-based wavefield computation}, pages 1644--1648.
\newblock 2022.
\newblock \doi{10.1190/image2022-3747298.1}.
\newblock URL
  \url{https://library.seg.org/doi/abs/10.1190/image2022-3747298.1}.

\bibitem[Cheng and Fu(2022)]{Cheng2022}
Yifan Cheng and Li-Yun Fu.
\newblock Nonlinear seismic inversion by physics-informed caianiello
  convolutional neural networks for overpressure prediction of source rocks in
  the offshore xihu depression, east china.
\newblock \emph{Journal of Petroleum Science and Engineering}, 215:\penalty0
  110654, August 2022.
\newblock \doi{10.1016/j.petrol.2022.110654}.
\newblock URL \url{https://doi.org/10.1016/j.petrol.2022.110654}.

\bibitem[Wei and Fu(2022)]{Wei2022}
Wei Wei and Li-Yun Fu.
\newblock Small-data-driven fast seismic simulations for complex media using
  physics-informed fourier neural operators.
\newblock \emph{{GEOPHYSICS}}, 87\penalty0 (6):\penalty0 T435--T446, November
  2022.
\newblock \doi{10.1190/geo2021-0573.1}.
\newblock URL \url{https://doi.org/10.1190/geo2021-0573.1}.

\bibitem[Li et~al.(2023)Li, Wang, Feng, Yang, and Lin]{Li2023}
Bian Li, Hanchen Wang, Shihang Feng, Xiu Yang, and Youzuo Lin.
\newblock Solving seismic wave equations on variable velocity models with
  fourier neural operator, 2023.

\bibitem[Yang et~al.(2021)Yang, Gao, Castellanos, Ross, Azizzadenesheli, and
  Clayton]{Yang2021}
Yan Yang, Angela~F. Gao, Jorge~C. Castellanos, Zachary~E. Ross, Kamyar
  Azizzadenesheli, and Robert~W. Clayton.
\newblock Seismic wave propagation and inversion with neural operators, 2021.

\bibitem[Dhara and Sen(2022)]{Dhara2022}
Arnab Dhara and Mrinal Sen.
\newblock Elastic-adjointnet: A physics guided deep autoencoder to overcome
  cross talk effects in multiparameter full waveform inversion.
\newblock volume 2022-August, pages 882--886. Society of Exploration
  Geophysicists, 8 2022.
\newblock \doi{10.1190/image2022-3745050.1}.

\bibitem[Huang and Alkhalifah(2022{\natexlab{a}})]{Huang2022b}
Xinquan Huang and Tariq Alkhalifah.
\newblock {PINNup}: Robust neural network wavefield solutions using frequency
  upscaling and neuron splitting.
\newblock \emph{Journal of Geophysical Research: Solid Earth}, 127\penalty0
  (6), jun 2022{\natexlab{a}}.
\newblock \doi{10.1029/2021jb023703}.
\newblock URL \url{https://doi.org/10.1029%2F2021jb023703}.

\bibitem[Alkhalifah et~al.(2021{\natexlab{a}})Alkhalifah, Song, Waheed, and
  Hao]{Alkhalifah2020b}
Tariq Alkhalifah, Chao Song, Umair~Bin Waheed, and Qi~Hao.
\newblock Wavefield solutions from machine learned functions constrained by the
  helmholtz equation.
\newblock \emph{Artificial Intelligence in Geosciences}, 2:\penalty0 11--19,
  December 2021{\natexlab{a}}.
\newblock \doi{10.1016/j.aiig.2021.08.002}.
\newblock URL \url{https://doi.org/10.1016/j.aiig.2021.08.002}.

\bibitem[Song et~al.(2021)Song, Alkhalifah, and Waheed]{Song2021}
Chao Song, Tariq Alkhalifah, and Umair~Bin Waheed.
\newblock A versatile framework to solve the helmholtz equation using
  physics-informed neural networks.
\newblock \emph{Geophysical Journal International}, 228\penalty0 (3):\penalty0
  1750--1762, October 2021.
\newblock \doi{10.1093/gji/ggab434}.
\newblock URL \url{https://doi.org/10.1093/gji/ggab434}.

\bibitem[Gou et~al.(2023)Gou, Zhang, Zhu, and Gao]{Gou2023}
Rongxi Gou, Yijie Zhang, Xueyu Zhu, and Jinghuai Gao.
\newblock Bayesian physics-informed neural networks for the subsurface
  tomography based on the eikonal equation.
\newblock \emph{IEEE Transactions on Geoscience and Remote Sensing}, pages
  1--1, 2023.
\newblock \doi{10.1109/TGRS.2023.3286438}.

\bibitem[Voytan and Sen(2020)]{Voytan2020}
Dimitri Voytan and Mrinal~K. Sen.
\newblock Wave propagation with physics informed neural networks.
\newblock volume 2020-October, pages 3477--3481. Society of Exploration
  Geophysicists, 2020.
\newblock \doi{10.1190/segam2020-3425406.1}.

\bibitem[Moseley et~al.(2020)Moseley, Markham, and Nissen-Meyer]{Moseley2020}
Ben Moseley, Andrew Markham, and Tarje Nissen-Meyer.
\newblock Solving the wave equation with physics-informed deep learning, 2020.

\bibitem[Kumar et~al.(2020)Kumar, Murali, and Priyadarshan]{Kumar2020}
Ashutosh Kumar, Amal Murali, and Amit Priyadarshan.
\newblock {Subsurface Velocity Profiling by Application of Physics Informed
  Neural Networks}.
\newblock volume Day 4 Thu, November 12, 2020 of \emph{Abu Dhabi International
  Petroleum Exhibition and Conference}, 11 2020.
\newblock \doi{10.2118/202766-MS}.
\newblock URL \url{https://doi.org/10.2118/202766-MS}.
\newblock D041S115R002.

\bibitem[Alkhalifah et~al.(2021{\natexlab{b}})Alkhalifah, Song, and
  Huang]{Alkhalifah2021}
Tariq Alkhalifah, Chao Song, and Xinquan Huang.
\newblock High-dimensional wavefield solutions based on neural network
  functions.
\newblock volume 2021-September, pages 2440--2444. Society of Exploration
  Geophysicists, 2021{\natexlab{b}}.
\newblock \doi{10.1190/segam2021-3584030.1}.

\bibitem[Huang and Alkhalifah(2022{\natexlab{b}})]{Huang2022}
Xinquan Huang and Tariq Alkhalifah.
\newblock \emph{Source location using physics-informed neural networks with
  hard constraints}, pages 1770--1774.
\newblock 2022{\natexlab{b}}.
\newblock \doi{10.1190/image2022-3738514.1}.
\newblock URL
  \url{https://library.seg.org/doi/abs/10.1190/image2022-3738514.1}.

\bibitem[Zhang et~al.(2023)Zhang, Zhu, and Gao]{YijieZhang2023}
Yijie Zhang, Xueyu Zhu, and Jinghuai Gao.
\newblock Seismic inversion based on acoustic wave equations using
  physics-informed neural network.
\newblock \emph{IEEE Transactions on Geoscience and Remote Sensing},
  61:\penalty0 1--11, 2023.
\newblock \doi{10.1109/TGRS.2023.3236973}.

\bibitem[Xu et~al.(2019)Xu, Li, and Chen]{Xu2019}
Yiran Xu, Jingye Li, and Xiaohong Chen.
\newblock \emph{Physics informed neural networks for velocity inversion}, pages
  2584--2588.
\newblock 2019.
\newblock \doi{10.1190/segam2019-3216823.1}.
\newblock URL
  \url{https://library.seg.org/doi/abs/10.1190/segam2019-3216823.1}.

\bibitem[Waheed et~al.(2021)Waheed, Alkhalifah, Haghighat, Song, and
  Virieux]{binWaheed2021}
Umair~Bin Waheed, Tariq Alkhalifah, Ehsan Haghighat, Chao Song, and Jean
  Virieux.
\newblock Pinntomo: Seismic tomography using physics-informed neural networks,
  2021.

\bibitem[Kharazmi et~al.(2021)Kharazmi, Wang, Fan, Rudy, Sapsis, Triantafyllou,
  and Karniadakis]{Kharazmi2021}
Ehsan Kharazmi, Zhicheng Wang, Dixia Fan, Samuel Rudy, Themis Sapsis, Michael~S
  Triantafyllou, and George~E Karniadakis.
\newblock From data to assessment models, demonstrated through a digital twin
  of marine risers, 2021.
\newblock URL
  \url{http://onepetro.org/OTCONF/proceedings-pdf/21OTC/3-21OTC/D031S035R003/2523337/otc-30985-ms.pdf}.

\bibitem[Kaneko et~al.(2023)Kaneko, Inoue, Nakagawa, Wada, Miyoshi, Abe,
  Kuroda, and Fujita]{Kaneko2023}
Tatsuya Kaneko, Tomoya Inoue, Yujin Nakagawa, Ryota Wada, Keisuke Miyoshi,
  Shungo Abe, Kouhei Kuroda, and Kazuhiro Fujita.
\newblock Hybrid approach using physical insights and data science for early
  stuck detection.
\newblock OTC, 4 2023.
\newblock \doi{10.4043/32532-MS}.
\newblock URL
  \url{https://onepetro.org/OTCONF/proceedings/23OTC/2-23OTC/D021S020R003/519301}.

\bibitem[Tang et~al.(2022)Tang, Liao, Yang, and Xie]{Tang2022}
Hesheng Tang, Yangyang Liao, Hu~Yang, and Liyu Xie.
\newblock A transfer learning-physics informed neural network ({TL}-{PINN}) for
  vortex-induced vibration.
\newblock \emph{Ocean Engineering}, 266:\penalty0 113101, December 2022.
\newblock \doi{10.1016/j.oceaneng.2022.113101}.
\newblock URL \url{https://doi.org/10.1016/j.oceaneng.2022.113101}.

\bibitem[Sheth et~al.(2022)Sheth, Indranil, Crispin, and José]{Prasham2022}
Prasham Sheth, Roychoudhury Indranil, Chatar Crispin, and Celaya José.
\newblock {A Hybrid Physics-Based and Machine-Learning Approach for Stick/Slip
  Prediction}.
\newblock volume Day 2 Wed, March 09, 2022 of \emph{SPE/IADC Drilling
  Conference and Exhibition}, 03 2022.
\newblock \doi{10.2118/208760-MS}.
\newblock URL \url{https://doi.org/10.2118/208760-MS}.
\newblock D022S005R001.

\bibitem[Zhang et~al.(2022{\natexlab{a}})Zhang, Sun, and Bai]{Zhang2022}
Tao Zhang, Shuyu Sun, and Hua Bai.
\newblock Thermodynamically-consistent flash calculation in energy industry:
  From iterative schemes to a unified thermodynamics-informed neural network.
\newblock \emph{International Journal of Energy Research}, 46\penalty0
  (11):\penalty0 15332--15346, June 2022{\natexlab{a}}.
\newblock \doi{10.1002/er.8234}.
\newblock URL \url{https://doi.org/10.1002/er.8234}.

\bibitem[Lv et~al.(2021)Lv, Cheng, Aghighi, Masoumi, and Roshan]{Lv2021}
Adelina Lv, Lei Cheng, Mohammad~Ali Aghighi, Hossein Masoumi, and Hamid Roshan.
\newblock A novel workflow based on physics-informed machine learning to
  determine the permeability profile of fractured coal seams using downhole
  geophysical logs.
\newblock \emph{Marine and Petroleum Geology}, 131:\penalty0 105171, September
  2021.
\newblock \doi{10.1016/j.marpetgeo.2021.105171}.
\newblock URL \url{https://doi.org/10.1016/j.marpetgeo.2021.105171}.

\bibitem[Yoga et~al.(2022)Yoga, Purswani, and Johns]{Yoga2022}
Hanif~Farrastama Yoga, Prakash Purswani, and Russell~Taylor Johns.
\newblock Predictive model for relative permeability using physics-based
  artificial neural networks.
\newblock In \emph{Day 2 Tue, April 26, 2022}. {SPE}, April 2022.
\newblock \doi{10.2118/209420-ms}.
\newblock URL \url{https://doi.org/10.2118/209420-ms}.

\bibitem[Zhang and Sun(2021)]{Zhang2021}
Tao Zhang and Shuyu Sun.
\newblock Thermodynamics-informed neural network ({TINN}) for phase equilibrium
  calculations considering capillary pressure.
\newblock \emph{Energies}, 14\penalty0 (22):\penalty0 7724, November 2021.
\newblock \doi{10.3390/en14227724}.
\newblock URL \url{https://doi.org/10.3390/en14227724}.

\bibitem[Hadjisotiriou et~al.(2023)Hadjisotiriou, Pour, and
  Voskov]{Hadjisotiriou2023}
George Hadjisotiriou, Kiarash~Mansour Pour, and Denis Voskov.
\newblock Application of deep neural networks to the operator space of
  nonlinear pde for physics-based proxy modelling.
\newblock Society of Petroleum Engineers, 2023.
\newblock ISBN 9781613998717.
\newblock \doi{10.2118/212217-MS}.

\bibitem[G\"{a}rttner et~al.(2023)G\"{a}rttner, Alpak, Meier, Ray, and
  Frank]{Grttner2023}
Stephan G\"{a}rttner, Faruk~O. Alpak, Andreas Meier, Nadja Ray, and Florian
  Frank.
\newblock Estimating permeability of 3d micro-{CT} images by physics-informed
  {CNNs} based on {DNS}.
\newblock \emph{Computational Geosciences}, 27\penalty0 (2):\penalty0 245--262,
  January 2023.
\newblock \doi{10.1007/s10596-022-10184-0}.
\newblock URL \url{https://doi.org/10.1007/s10596-022-10184-0}.

\bibitem[Boateng et~al.(2017)Boateng, Fu, {Yu}, and {Xizhu}]{Boateng2017}
Cyril~D. Boateng, Li-Yun Fu, Wu~{Yu}, and Guan {Xizhu}.
\newblock {Porosity inversion by Caianiello neural networks with
  Levenberg-Marquardt optimization}.
\newblock \emph{Interpretation}, 5\penalty0 (3):\penalty0 SL33--SL42, August
  2017.
\newblock \doi{10.1190/INT-2016-0119.1}.

\bibitem[Tembely et~al.(2021)Tembely, AlSumaiti, and Alameri]{Tembely2021}
Moussa Tembely, Ali~M. AlSumaiti, and Waleed~S. Alameri.
\newblock Machine and deep learning for estimating the permeability of complex
  carbonate rock from x-ray micro-computed tomography.
\newblock \emph{Energy Reports}, 7:\penalty0 1460--1472, November 2021.
\newblock \doi{10.1016/j.egyr.2021.02.065}.
\newblock URL \url{https://doi.org/10.1016/j.egyr.2021.02.065}.

\bibitem[Wu et~al.(2018)Wu, Yin, and Xiao]{Wu2018}
Jinlong Wu, Xiaolong Yin, and Heng Xiao.
\newblock Seeing permeability from images: fast prediction with convolutional
  neural networks.
\newblock \emph{Science Bulletin}, 63\penalty0 (18):\penalty0 1215--1222,
  September 2018.
\newblock \doi{10.1016/j.scib.2018.08.006}.
\newblock URL \url{https://doi.org/10.1016/j.scib.2018.08.006}.

\bibitem[Ihunde and Olorode(2021)]{Ihunde2021}
Thelma~Anizia Ihunde and Olufemi Olorode.
\newblock Application of physics informed neural networks to compositional
  modelling.
\newblock In \emph{Proceedings of the 2021 Asia Pacific Unconventional
  Resources Technology Conference}. Unconventional Resources Technology
  Conference, 2021.
\newblock \doi{10.15530/ap-urtec-2021-208310}.
\newblock URL \url{https://doi.org/10.15530/ap-urtec-2021-208310}.

\bibitem[Behl and Tyagi(2023)]{Behl2023}
M~V Behl and M~Tyagi.
\newblock Data-driven reduced-order models for volve field using reservoir
  simulation and physics-informed machine learning techniques.
\newblock 2023.
\newblock \doi{10.2118/214288-PA/3065651/spe-214288-pa.pdf}.
\newblock URL
  \url{http://onepetro.org/REE/article-pdf/doi/10.2118/214288-PA/3065651/spe-214288-pa.pdf}.

\bibitem[Wu and Sun(2023)]{Wu2023}
Yuanqing Wu and Shuyu Sun.
\newblock {Removing the performance bottleneck of pressure–temperature flash
  calculations during both the online and offline stages by using
  physics-informed neural networks}.
\newblock \emph{Physics of Fluids}, 35\penalty0 (4), 04 2023.
\newblock ISSN 1070-6631.
\newblock \doi{10.1063/5.0150341}.
\newblock URL \url{https://doi.org/10.1063/5.0150341}.
\newblock 043326.

\bibitem[Deng and Pan(2021)]{Deng2021}
Lichi Deng and Yuewei Pan.
\newblock Application of physics-informed neural networks for self-similar and
  transient solutions of spontaneous imbibition.
\newblock \emph{Journal of Petroleum Science and Engineering}, 203:\penalty0
  108644, August 2021.
\newblock \doi{10.1016/j.petrol.2021.108644}.
\newblock URL \url{https://doi.org/10.1016/j.petrol.2021.108644}.

\bibitem[Abbasi and Østebø Andersen(2023)]{Abbasi2023}
Jassem Abbasi and Pål Østebø Andersen.
\newblock Simulation and prediction of countercurrent spontaneous imbibition at
  early and late times using physics-informed neural networks.
\newblock In \emph{Day 3 Wed, June 07, 2023}. {SPE}, June 2023.
\newblock \doi{10.2118/214433-ms}.
\newblock URL \url{https://doi.org/10.2118/214433-ms}.

\bibitem[Wang et~al.(2022)Wang, Zhang, and Wu]{Wang2022}
Shihao Wang, Yanbin Zhang, and Yu-Shu Wu.
\newblock Deep learning accelerated hydraulic-mechanical simulation with prior
  smoothness constraints for heterogeneous petroleum reservoirs, 2022.
\newblock URL
  \url{http://onepetro.org/SJ/article-pdf/27/05/2689/3014841/spe-201430-pa.pdf}.

\bibitem[Chen et~al.(2023)Chen, Luo, Wang, and Nasrabadi]{Chen2023}
Fangxuan Chen, Sheng Luo, Shihao Wang, and Hadi Nasrabadi.
\newblock A novel machine-learning assisted phase-equilibrium calculation model
  for liquid-rich shale reservoirs.
\newblock Society of Petroleum Engineers, 2023.
\newblock ISBN 9781613998717.
\newblock \doi{10.2118/212193-MS}.

\bibitem[Haghighat et~al.(2022)Haghighat, Amini, and Juanes]{Haghighat2022}
Ehsan Haghighat, Danial Amini, and Ruben Juanes.
\newblock Physics-informed neural network simulation of multiphase
  poroelasticity using stress-split sequential training.
\newblock \emph{Computer Methods in Applied Mechanics and Engineering},
  397:\penalty0 115141, July 2022.
\newblock \doi{10.1016/j.cma.2022.115141}.
\newblock URL \url{https://doi.org/10.1016/j.cma.2022.115141}.

\bibitem[Diab et~al.(2022)Diab, Chaabi, Zhang, Arif, Alkobaisi, and
  Kobaisi]{Diab2022}
Waleed Diab, Omar Chaabi, Wenjuan Zhang, Muhammad Arif, Shayma Alkobaisi, and
  Mohammed~Al Kobaisi.
\newblock Data-free and data-efficient physics-informed neural network
  approaches to solve the buckley{\textendash}leverett problem.
\newblock \emph{Energies}, 15\penalty0 (21):\penalty0 7864, October 2022.
\newblock \doi{10.3390/en15217864}.
\newblock URL \url{https://doi.org/10.3390/en15217864}.

\bibitem[Torrado et~al.(2021)Torrado, Ruiz, Cueto{-}Felgueroso, Green, Friesen,
  Matringe, and Togelius]{Torrado2021}
Ruben~Rodriguez Torrado, Pablo Ruiz, Luis Cueto{-}Felgueroso, Michael~Cerny
  Green, Tyler Friesen, Sebastien Matringe, and Julian Togelius.
\newblock Physics-informed attention-based neural network for solving
  non-linear partial differential equations.
\newblock \emph{CoRR}, abs/2105.07898, 2021.
\newblock URL \url{https://arxiv.org/abs/2105.07898}.

\bibitem[Diab et~al.(2023)Diab, Chaabi, Alkobaisi, Awotunde, and
  Kobaisi]{Diab2023}
Waleed Diab, Omar Chaabi, Shayma Alkobaisi, Abeeb Awotunde, and Mohammed~Al
  Kobaisi.
\newblock Learning generic solutions for multiphase transport in porous media
  via the flux functions operator, 2023.

\bibitem[Maucec and Jalali(2022)]{Maucec2022}
Marko Maucec and Ridwan Jalali.
\newblock Geodin-geoscience-based deep interaction networks for predicting flow
  dynamics in reservoir simulation models, 2022.
\newblock URL
  \url{http://onepetro.org/SJ/article-pdf/27/03/1671/2726009/spe-203952-pa.pdf}.

\bibitem[Battaglia et~al.(2016)Battaglia, Pascanu, Lai, Jimenez~Rezende, and
  kavukcuoglu]{Battaglia2016}
Peter Battaglia, Razvan Pascanu, Matthew Lai, Danilo Jimenez~Rezende, and koray
  kavukcuoglu.
\newblock Interaction networks for learning about objects, relations and
  physics.
\newblock In D.~Lee, M.~Sugiyama, U.~Luxburg, I.~Guyon, and R.~Garnett,
  editors, \emph{Advances in Neural Information Processing Systems}, volume~29.
  Curran Associates, Inc., 2016.
\newblock URL
  \url{https://proceedings.neurips.cc/paper_files/paper/2016/file/3147da8ab4a0437c15ef51a5cc7f2dc4-Paper.pdf}.

\bibitem[Zhang et~al.(2022{\natexlab{b}})Zhang, Zuo, Zhao, Ma, Gu, Wang, Yang,
  Yao, and Yao]{KaiZhang2022}
Kai Zhang, Yuande Zuo, Hanjun Zhao, Xiaopeng Ma, Jianwei Gu, Jian Wang, Yongfei
  Yang, Chuanjin Yao, and Jun Yao.
\newblock Fourier neural operator for solving subsurface oil/water two-phase
  flow partial differential equation, 2022{\natexlab{b}}.
\newblock URL
  \url{http://onepetro.org/SJ/article-pdf/27/03/1815/2726200/spe-209223-pa.pdf}.

\bibitem[Wen et~al.(2022)Wen, Li, Azizzadenesheli, Anandkumar, and
  Benson]{Wen2022}
Gege Wen, Zongyi Li, Kamyar Azizzadenesheli, Anima Anandkumar, and Sally~M.
  Benson.
\newblock U-fno -- an enhanced fourier neural operator-based deep-learning
  model for multiphase flow, 2022.

\bibitem[Zhang(2022)]{ZhaoZhang2022}
Zhao Zhang.
\newblock A physics-informed deep convolutional neural network for simulating
  and predicting transient darcy flows in heterogeneous reservoirs without
  labeled data.
\newblock \emph{Journal of Petroleum Science and Engineering}, 211:\penalty0
  110179, April 2022.
\newblock \doi{10.1016/j.petrol.2022.110179}.
\newblock URL \url{https://doi.org/10.1016/j.petrol.2022.110179}.

\bibitem[Fuks and Tchelepi(2020)]{Fuks2020}
Olga Fuks and Hamdi~A Tchelepi.
\newblock Limitations of physics informed machine learning for nonlinear
  two-phase transport in porous media, 2020.
\newblock URL \url{www.begellhouse.com}.

\bibitem[Magzymov and Dindoruk(2021)]{Magzymov2021}
Daulet Magzymov and Birol Dindoruk.
\newblock Evaluation of machine learning methodologies using simple physics
  based conceptual models for flow in porous media.
\newblock pages 21--23, 2021.
\newblock URL
  \url{http://onepetro.org/SPEATCE/proceedings-pdf/21ATCE/2-21ATCE/D021S038R004/2498940/spe-206359-ms.pdf}.

\bibitem[Gasmi and Tchelepi(2021)]{Gasmi2021}
Cedric~Fraces Gasmi and Hamdi Tchelepi.
\newblock Physics informed deep learning for flow and transport in porous
  media.
\newblock 2021.

\bibitem[Alhubail et~al.(2022)Alhubail, He, AlSinan, Kwak, and
  Hoteit]{Alhubail2022}
Ali Alhubail, Xupeng He, Marwa AlSinan, Hyung Kwak, and Hussein Hoteit.
\newblock Extended physics-informed neural networks for solving fluid flow
  problems in highly heterogeneous media.
\newblock {IPTC}, 2022.
\newblock \doi{10.2523/iptc-22163-ms}.
\newblock URL \url{https://doi.org/10.2523/iptc-22163-ms}.

\bibitem[Sambo and Feng(2021)]{Sambo2021}
Chico Sambo and Yin Feng.
\newblock Physics inspired machine learning for solving fluid flow in porous
  media: A novel computational algorithm for reservoir simulation, 2021.
\newblock URL
  \url{http://onepetro.org/spersc/proceedings-pdf/21RSC/1-21RSC/D011S008R007/2507961/spe-203917-ms.pdf}.

\bibitem[Shen et~al.(2022)Shen, Li, Zha, Li, and Liu]{Shen2022}
Luhang Shen, Daolun Li, Wenshu Zha, Xiang Li, and Xuliang Liu.
\newblock Surrogate modeling for porous flow using deep neural networks.
\newblock \emph{Journal of Petroleum Science and Engineering}, 213:\penalty0
  110460, June 2022.
\newblock \doi{10.1016/j.petrol.2022.110460}.
\newblock URL \url{https://doi.org/10.1016/j.petrol.2022.110460}.

\bibitem[Almajid and Abu-Alsaud(2020)]{Almajid2020}
Muhammad~Majid Almajid and Moataz~Omar Abu-Alsaud.
\newblock {Prediction of Fluid Flow in Porous Media using Physics Informed
  Neural Networks}.
\newblock volume Day 2 Tue, November 10, 2020 of \emph{Abu Dhabi International
  Petroleum Exhibition and Conference}, 11 2020.
\newblock \doi{10.2118/203033-MS}.
\newblock URL \url{https://doi.org/10.2118/203033-MS}.
\newblock D021S050R004.

\bibitem[Coutinho et~al.(2023)Coutinho, Dall{\textquotesingle}Aqua, McClenny,
  Zhong, Braga-Neto, and Gildin]{Coutinho2023}
Emilio Jose~Rocha Coutinho, Marcelo Dall{\textquotesingle}Aqua, Levi McClenny,
  Ming Zhong, Ulisses Braga-Neto, and Eduardo Gildin.
\newblock Physics-informed neural networks with adaptive localized artificial
  viscosity.
\newblock \emph{Journal of Computational Physics}, 489:\penalty0 112265,
  September 2023.
\newblock \doi{10.1016/j.jcp.2023.112265}.
\newblock URL \url{https://doi.org/10.1016/j.jcp.2023.112265}.

\bibitem[He et~al.(2020)He, Barajas-Solano, Tartakovsky, and
  Tartakovsky]{He2020}
QiZhi He, David Barajas-Solano, Guzel Tartakovsky, and Alexandre~M.
  Tartakovsky.
\newblock Physics-informed neural networks for multiphysics data assimilation
  with application to subsurface transport.
\newblock \emph{Advances in Water Resources}, 141:\penalty0 103610, July 2020.
\newblock \doi{10.1016/j.advwatres.2020.103610}.
\newblock URL \url{https://doi.org/10.1016/j.advwatres.2020.103610}.

\bibitem[Manasipov et~al.(2023)Manasipov, Nikolaev, Didenko, Abdalla, and
  Stundner]{Manasipov2023}
Roman Manasipov, Denis Nikolaev, Dmitrii Didenko, Ramez Abdalla, and Michael
  Stundner.
\newblock Physics informed machine learning for production forecast.
\newblock Society of Petroleum Engineers, 2023.
\newblock ISBN 9781613999738.
\newblock \doi{10.2118/212666-MS}.

\bibitem[Nagao et~al.(2023)Nagao, Datta-Gupta, Onishi, and Sankaran]{Nagao2023}
Masahiro Nagao, Akhil Datta-Gupta, Tsubasa Onishi, and Sathish Sankaran.
\newblock Reservoir connectivity identification and robust production
  forecasting using physics informed machine learning.
\newblock Society of Petroleum Engineers, 2023.
\newblock ISBN 9781613998717.
\newblock \doi{10.2118/212201-MS}.

\bibitem[Liu et~al.(2023{\natexlab{a}})Liu, Jing, and Pan]{Liu2023}
Jianqiao Liu, Hongbin Jing, and Huanquan Pan.
\newblock New fast simulation of 4d (x, y, z, t) co2 eor by fourier neural
  operator based deep learning method.
\newblock Society of Petroleum Engineers, 2023{\natexlab{a}}.
\newblock ISBN 9781613998717.
\newblock \doi{10.2118/212236-MS}.

\bibitem[Darabi et~al.(2022)Darabi, Kianinejad, and Salehi]{Darabi2022}
Hamed Darabi, Amir Kianinejad, and Amir Salehi.
\newblock {Physics-Informed Spatio-Temporal Graph Neural Network for Waterflood
  Management}.
\newblock volume Day 1 Mon, October 31, 2022 of \emph{Abu Dhabi International
  Petroleum Exhibition and Conference}, 10 2022.
\newblock \doi{10.2118/211284-MS}.
\newblock URL \url{https://doi.org/10.2118/211284-MS}.
\newblock D011S017R004.

\bibitem[Sarma et~al.(2017)Sarma, Kyriacou, Henning, Orland, Thakur, and
  Sloss]{Sarma2017}
Pallav Sarma, Stylianos Kyriacou, Mark Henning, Paul Orland, Ganesh Thakur, and
  Dakin Sloss.
\newblock Redistribution of steam injection in heavy oil reservoir management
  to improve {EOR} economics, powered by a unique integration of reservoir
  physics and machine learning.
\newblock In \emph{Day 2 Thu, May 18, 2017}. {SPE}, May 2017.
\newblock \doi{10.2118/185507-ms}.
\newblock URL \url{https://doi.org/10.2118/185507-ms}.

\bibitem[Gladchenko et~al.(2023)Gladchenko, Illarionov, Orlov, and
  Koroteev]{Gladchenko2023}
Elizaveta~S. Gladchenko, Egor~A. Illarionov, Denis~M. Orlov, and Dmitry~A.
  Koroteev.
\newblock Physics-informed neural networks and capacitance-resistance model:
  Fast and accurate oil and water production forecast using end-to-end
  architecture.
\newblock In \emph{Day 2 Wed, January 18, 2023}. {SPE}, January 2023.
\newblock \doi{10.2118/214461-ms}.
\newblock URL \url{https://doi.org/10.2118/214461-ms}.

\bibitem[Maniglio et~al.(2021)Maniglio, Fighera, Dovera, and
  Stabile]{Maniglio2021}
Marco Maniglio, Giorgio Fighera, Laura Dovera, and Carlo~Cristiano Stabile.
\newblock {Physics Informed Neural Networks Based on a Capacitance Resistance
  Model for Reservoirs Under Water Flooding Conditions}.
\newblock volume Day 2 Tue, November 16, 2021 of \emph{Abu Dhabi International
  Petroleum Exhibition and Conference}, 11 2021.
\newblock \doi{10.2118/207800-MS}.
\newblock URL \url{https://doi.org/10.2118/207800-MS}.
\newblock D021S027R001.

\bibitem[Molinari and Sankaran(2021)]{Molinari2021}
Diego Molinari and Sathish Sankaran.
\newblock Merging physics and data-driven methods for field-wide bottomhole
  pressure estimation in unconventional wells.
\newblock American Association of Petroleum Geologists AAPG/Datapages, 8 2021.
\newblock \doi{10.15530/urtec-2021-5261}.

\bibitem[Harp et~al.(2021)Harp, O'Malley, Yan, and Pawar]{Harp2021}
Dylan~Robert Harp, Dan O'Malley, Bicheng Yan, and Rajesh Pawar.
\newblock On the feasibility of using physics-informed machine learning for
  underground reservoir pressure management.
\newblock \emph{Expert Systems with Applications}, 178:\penalty0 115006,
  September 2021.
\newblock \doi{10.1016/j.eswa.2021.115006}.
\newblock URL \url{https://doi.org/10.1016/j.eswa.2021.115006}.

\bibitem[Staff et~al.(2020)Staff, Zarruk, Hatleskog, Stavland, McNulty, and
  Ibarra]{Staff2020}
Gunnar Staff, Gustavo Zarruk, Johan Hatleskog, Simon Stavland, Henry McNulty,
  and Roberto Ibarra.
\newblock Physics guided machine learning significantly improves outcomes for
  data-based production optimization, 2020.
\newblock URL
  \url{http://onepetro.org/SPEADIP/proceedings-pdf/20ADIP/1-20ADIP/D011S019R003/2375793/spe-202657-ms.pdf}.

\bibitem[Razak et~al.(2021)Razak, Cornelio, Cho, Liu, Vaidya, and
  Jafarpour]{Razak2021c}
Syamil~Mohd Razak, Jodel Cornelio, Young Cho, Hui-Hai Liu, Ravimadhav Vaidya,
  and Behnam Jafarpour.
\newblock A physics-guided deep learning predictive model for robust production
  forecasting and diagnostics in unconventional wells.
\newblock 7 2021.
\newblock \doi{10.15530/URTEC-2021-5059}.
\newblock URL \url{/URTECONF/proceedings-abstract/21URTC/2-21URTC/465267}.

\bibitem[Razak et~al.(2022)Razak, Cornelio, Cho, Liu, Vaidya, and
  Jafarpour]{Razak2022}
Syamil~Mohd Razak, Jodel Cornelio, Young Cho, Hui-Hai Liu, Ravimadhav Vaidya,
  and Behnam Jafarpour.
\newblock Embedding physical flow functions into deep learning predictive
  models for improved production forecasting.
\newblock American Association of Petroleum Geologists AAPG/Datapages, 7 2022.
\newblock \doi{10.15530/urtec-2022-3702606}.

\bibitem[Busby(2020)]{Busby2020}
D.~Busby.
\newblock Deep-{DCA} a new approach for well hydrocarbon production
  forecasting.
\newblock In \emph{{ECMOR} {XVII}}. European Association of Geoscientists {\&}
  Engineers, 2020.
\newblock \doi{10.3997/2214-4609.202035124}.
\newblock URL \url{https://doi.org/10.3997/2214-4609.202035124}.

\bibitem[Franklin et~al.(2022)Franklin, Souza, Fontes, and
  Martins]{Franklin2022}
Taniel~S. Franklin, Leonardo~S. Souza, Raony~M. Fontes, and M{\'{a}}rcio~A.F.
  Martins.
\newblock A physics-informed neural networks ({PINN}) oriented approach to flow
  metering in oil wells: an {ESP} lifted oil well system as a case study.
\newblock \emph{Digital Chemical Engineering}, 5:\penalty0 100056, December
  2022.
\newblock \doi{10.1016/j.dche.2022.100056}.
\newblock URL \url{https://doi.org/10.1016/j.dche.2022.100056}.

\bibitem[Liu et~al.(2023{\natexlab{b}})Liu, Vashisth, Grana, and
  Mukerji]{MingliangLiu2023}
Mingliang Liu, Divakar Vashisth, Dario Grana, and Tapan Mukerji.
\newblock Joint inversion of geophysical data for geologic carbon sequestration
  monitoring: A differentiable physics-informed neural network model.
\newblock \emph{Journal of Geophysical Research: Solid Earth}, 128\penalty0
  (3), March 2023{\natexlab{b}}.
\newblock \doi{10.1029/2022jb025372}.
\newblock URL \url{https://doi.org/10.1029/2022jb025372}.

\bibitem[Tariq et~al.(2023)Tariq, Yan, and Sun]{Tariq2023}
Zeeshan Tariq, Bicheng Yan, and Shuyu Sun.
\newblock Physics informed surrogate model development in predicting dynamic
  temporal and spatial variations during {CO}2 injection into deep saline
  aquifers.
\newblock In \emph{Day 2 Wed, January 25, 2023}. {SPE}, January 2023.
\newblock \doi{10.2118/212693-ms}.
\newblock URL \url{https://doi.org/10.2118/212693-ms}.

\bibitem[Jiang et~al.(2023)Jiang, Zhu, Li, Li, Yuan, and Lu]{Jiang2023}
Zhongyi Jiang, Min Zhu, Dongzhuo Li, Qiuzi Li, Yanhua~O. Yuan, and Lu~Lu.
\newblock Fourier-mionet: Fourier-enhanced multiple-input neural operators for
  multiphase modeling of geological carbon sequestration, 2023.

\bibitem[Yan et~al.(2022{\natexlab{a}})Yan, Harp, Chen, and Pawar]{Yan2022}
Bicheng Yan, Dylan~Robert Harp, Bailian Chen, and Rajesh Pawar.
\newblock A physics-constrained deep learning model for simulating multiphase
  flow in 3d heterogeneous porous media.
\newblock \emph{Fuel}, 313:\penalty0 122693, April 2022{\natexlab{a}}.
\newblock \doi{10.1016/j.fuel.2021.122693}.
\newblock URL \url{https://doi.org/10.1016/j.fuel.2021.122693}.

\bibitem[Du et~al.(2023)Du, Zhao, Cheng, Yan, and He]{HonghuiDu2023}
Honghui Du, Ze~Zhao, Haojia Cheng, Jinhui Yan, and QiZhi He.
\newblock Modeling density-driven flow in porous media by physics-informed
  neural networks for {CO}2 sequestration.
\newblock \emph{Computers and Geotechnics}, 159:\penalty0 105433, July 2023.
\newblock \doi{10.1016/j.compgeo.2023.105433}.
\newblock URL \url{https://doi.org/10.1016/j.compgeo.2023.105433}.

\bibitem[Shokouhi et~al.(2021)Shokouhi, Kumar, Prathipati, Hosseini, Giles, and
  Kifer]{Shokouhi2021}
Parisa Shokouhi, Vikas Kumar, Sumedha Prathipati, Seyyed~A. Hosseini, Clyde~Lee
  Giles, and Daniel Kifer.
\newblock Physics-informed deep learning for prediction of {CO}2 storage site
  response.
\newblock \emph{Journal of Contaminant Hydrology}, 241:\penalty0 103835, August
  2021.
\newblock \doi{10.1016/j.jconhyd.2021.103835}.
\newblock URL \url{https://doi.org/10.1016/j.jconhyd.2021.103835}.

\bibitem[Yan et~al.(2022{\natexlab{b}})Yan, Harp, Chen, Hoteit, and
  Pawar]{Yan2022b}
Bicheng Yan, Dylan~Robert Harp, Bailian Chen, Hussein Hoteit, and Rajesh~J.
  Pawar.
\newblock A gradient-based deep neural network model for simulating multiphase
  flow in porous media.
\newblock \emph{Journal of Computational Physics}, 463:\penalty0 111277, August
  2022{\natexlab{b}}.
\newblock \doi{10.1016/j.jcp.2022.111277}.
\newblock URL \url{https://doi.org/10.1016/j.jcp.2022.111277}.

\bibitem[Sitzmann et~al.(2020)Sitzmann, Martel, Bergman, Lindell, and
  Wetzstein]{Sitzmann2020}
Vincent Sitzmann, Julien N.~P. Martel, Alexander~W. Bergman, David~B. Lindell,
  and Gordon Wetzstein.
\newblock Implicit neural representations with periodic activation functions.
\newblock \emph{CoRR}, abs/2006.09661, 2020.
\newblock URL \url{https://arxiv.org/abs/2006.09661}.

\bibitem[Grubas et~al.(2021)Grubas, Yaskevich, and Duchkov]{Grubas2021}
S.~Grubas, S.~Yaskevich, and A.~Duchkov.
\newblock Localization of microseismic events using the physics-informed
  neural-network for traveltime computation.
\newblock In \emph{82nd {EAGE} Annual Conference \& Exhibition}. European
  Association of Geoscientists {\&} Engineers, 2021.
\newblock \doi{10.3997/2214-4609.202113191}.
\newblock URL \url{https://doi.org/10.3997/2214-4609.202113191}.

\bibitem[Li et~al.(2021{\natexlab{a}})Li, Kovachki, Azizzadenesheli, Liu,
  Bhattacharya, Stuart, and Anandkumar]{Li2021b}
Zongyi Li, Nikola Kovachki, Kamyar Azizzadenesheli, Burigede Liu, Kaushik
  Bhattacharya, Andrew Stuart, and Anima Anandkumar.
\newblock Fourier neural operator for parametric partial differential
  equations, 2021{\natexlab{a}}.

\bibitem[Li et~al.(2021{\natexlab{b}})Li, Kovachki, Azizzadenesheli, Liu,
  Bhattacharya, Stuart, and Anandkumar]{Li2021}
Zongyi Li, Nikola Kovachki, Kamyar Azizzadenesheli, Burigede Liu, Kaushik
  Bhattacharya, Andrew Stuart, and Anima Anandkumar.
\newblock Fourier neural operator for parametric partial differential
  equations, 2021{\natexlab{b}}.

\bibitem[Operto et~al.(2013)Operto, Gholami, Prieux, Ribodetti, Brossier,
  Metivier, and Virieux]{Operto2013}
S.~Operto, Y.~Gholami, V.~Prieux, A.~Ribodetti, R.~Brossier, L.~Metivier, and
  J.~Virieux.
\newblock A guided tour of multiparameter full-waveform inversion with
  multicomponent data: From theory to practice.
\newblock \emph{The Leading Edge}, 32\penalty0 (9):\penalty0 1040--1054, 2013.
\newblock \doi{10.1190/tle32091040.1}.
\newblock URL \url{https://doi.org/10.1190/tle32091040.1}.

\bibitem[Rahaman et~al.(2019)Rahaman, Baratin, Arpit, Draxler, Lin, Hamprecht,
  Bengio, and Courville]{Rahaman2019}
Nasim Rahaman, Aristide Baratin, Devansh Arpit, Felix Draxler, Min Lin, Fred~A.
  Hamprecht, Yoshua Bengio, and Aaron Courville.
\newblock On the spectral bias of neural networks, 2019.

\bibitem[Jagtap et~al.(2021)Jagtap, Shin, Kawaguchi, and
  Karniadakis]{Jagtap2021}
Ameya~D. Jagtap, Yeonjong Shin, Kenji Kawaguchi, and George~Em Karniadakis.
\newblock Deep kronecker neural networks: A general framework for neural
  networks with adaptive activation functions, 2021.

\bibitem[McCulloch and Pitts(1943)]{McCulloch1943}
Warren~S. McCulloch and Walter Pitts.
\newblock A logical calculus of the ideas immanent in nervous activity.
\newblock \emph{The Bulletin of Mathematical Biophysics}, 5\penalty0
  (4):\penalty0 115 – 133, 1943.
\newblock \doi{10.1007/BF02478259}.
\newblock URL
  \url{https://www.scopus.com/inward/record.uri?eid=2-s2.0-51249194645&doi=10.1007%2fBF02478259&partnerID=40&md5=edb67afceee33d22eaabbf1f8c1dca90}.

\bibitem[Caianiello(1961)]{Caianiello1961}
E.R. Caianiello.
\newblock Outline of a theory of thought-processes and thinking machines.
\newblock \emph{Journal of Theoretical Biology}, 1\penalty0 (2):\penalty0
  204--235, 1961.
\newblock ISSN 0022-5193.
\newblock \doi{https://doi.org/10.1016/0022-5193(61)90046-7}.
\newblock URL
  \url{https://www.sciencedirect.com/science/article/pii/0022519361900467}.

\bibitem[Fu(2001)]{Fu2001}
Li-Yun Fu.
\newblock Chapter 12 caianiello neural network method for geophysical inverse
  problems.
\newblock In \emph{Computational neural networks for geophysical data
  processing}, pages 187--215. Elsevier, 2001.
\newblock \doi{10.1016/s0950-1401(01)80026-4}.
\newblock URL \url{https://doi.org/10.1016/s0950-1401(01)80026-4}.

\bibitem[Brandolin et~al.(2022)Brandolin, Ravasi, and
  Alkhalifah]{Brandolin2022}
Francesco Brandolin, Matteo Ravasi, and Tariq Alkhalifah.
\newblock \emph{PWD-PINN: Slope-assisted seismic interpolation with
  physics-informed neural networks}, pages 2646--2650.
\newblock 2022.
\newblock \doi{10.1190/image2022-3742422.1}.
\newblock URL
  \url{https://library.seg.org/doi/abs/10.1190/image2022-3742422.1}.

\bibitem[Rudat and Dashevskiy(2011)]{Rudat2011}
Jens Rudat and Dmitriy Dashevskiy.
\newblock {Development of an Innovative Model-Based Stick/Slip Control System}.
\newblock volume All Days of \emph{SPE/IADC Drilling Conference and
  Exhibition}, 03 2011.
\newblock \doi{10.2118/139996-MS}.
\newblock URL \url{https://doi.org/10.2118/139996-MS}.
\newblock SPE-139996-MS.

\bibitem[Jan et~al.(2022)Jan, Mahfoudh, Draškovic, Jeong, and Yu]{Jan2022}
A.~Jan, F.~Mahfoudh, G.~Draškovic, C.~Jeong, and Y.~Yu.
\newblock Multitasking physics-informed neural network for drillstring washout
  detection.
\newblock In \emph{83nd {EAGE} Annual Conference \& Exhibition}. European
  Association of Geoscientists \& Engineers, 2022.
\newblock \doi{10.3997/2214-4609.202210607}.
\newblock URL \url{https://doi.org/10.3997/2214-4609.202210607}.

\bibitem[Mellal et~al.(2023)Mellal, Rasouli, Dehdouh, Latrach, Abdelhamid,
  Malki, and Bakelli]{Mellal2023}
Ilyas Mellal, Vamegh Rasouli, Abdesselem Dehdouh, Abdeldjalil Latrach, Cilia
  Abdelhamid, Mohamed Malki, and Omar Bakelli.
\newblock Formation evaluation challenges of tight and shale reservoirs. a case
  study of the bakken petroleum system.
\newblock 06 2023.
\newblock \doi{10.56952/arma-2023-0894}.

\bibitem[Fraces and Tchelepi(2023)]{Gasmi2023}
Cedric~G. Fraces and Hamdi Tchelepi.
\newblock Uncertainty quantification for transport in porous media using
  parameterized physics informed neural networks.
\newblock In \emph{Day 1 Tue, March 28, 2023}. {SPE}, March 2023.
\newblock \doi{10.2118/212255-ms}.
\newblock URL \url{https://doi.org/10.2118/212255-ms}.

\bibitem[Strelow et~al.(2023)Strelow, Gerisch, Lang, and Pfetsch]{Strelow2023}
Erik~Laurin Strelow, Alf Gerisch, Jens Lang, and Marc~E. Pfetsch.
\newblock Physics informed neural networks: A case study for gas transport
  problems.
\newblock \emph{Journal of Computational Physics}, 481:\penalty0 112041, 5
  2023.
\newblock ISSN 00219991.
\newblock \doi{10.1016/j.jcp.2023.112041}.

\bibitem[Dana et~al.(2021)Dana, Jammoul, and Wheeler]{Dana2021}
Saumik Dana, Mohamad Jammoul, and Mary~F. Wheeler.
\newblock Performance studies of the fixed stress split algorithm for
  immiscible two-phase flow coupled with linear poromechanics.
\newblock \emph{Computational Geosciences}, 26\penalty0 (1):\penalty0 13--27,
  October 2021.
\newblock \doi{10.1007/s10596-021-10110-w}.
\newblock URL \url{https://doi.org/10.1007/s10596-021-10110-w}.

\bibitem[D.~Jagtap and Em~Karniadakis(2020)]{Jagtap2020b}
Ameya D.~Jagtap and George Em~Karniadakis.
\newblock Extended physics-informed neural networks (xpinns): A generalized
  space-time domain decomposition based deep learning framework for nonlinear
  partial differential equations.
\newblock \emph{Communications in Computational Physics}, 28\penalty0
  (5):\penalty0 2002--2041, 2020.
\newblock ISSN 1991-7120.
\newblock \doi{https://doi.org/10.4208/cicp.OA-2020-0164}.
\newblock URL \url{http://global-sci.org/intro/article_detail/cicp/18403.html}.

\bibitem[Holanda et~al.(2018)Holanda, Gildin, Jensen, Lake, and
  Kabir]{Holanda2018}
Rafael Holanda, Eduardo Gildin, Jerry Jensen, Larry Lake, and C.~Kabir.
\newblock A state-of-the-art literature review on capacitance resistance models
  for reservoir characterization and performance forecasting.
\newblock \emph{Energies}, 11\penalty0 (12):\penalty0 3368, December 2018.
\newblock \doi{10.3390/en11123368}.
\newblock URL \url{https://doi.org/10.3390/en11123368}.

\bibitem[Zhao et~al.(2018)Zhao, Sarma, and Corp]{Zhao2018}
Yong Zhao, Pallav Sarma, and Tachyus Corp.
\newblock A benchmarking study of a novel data physics technology for
  steamflood and sagd modeling: Comparison to conventional reservoir
  simulation, 2018.
\newblock URL
  \url{http://onepetro.org/SPECHOC/proceedings-pdf/18CHOC/2-18CHOC/D021S006R001/1228328/spe-189772-ms.pdf/1}.

\bibitem[Shawaf et~al.(2023)Shawaf, Rasouli, and Dehdouh]{Shawaf2023}
Ali Shawaf, Vamegh Rasouli, and Abdesselem Dehdouh.
\newblock The impact of formation anisotropy and stresses on fractural
  geometry{\textemdash}a case study in jafurah's tuwaiq mountain formation
  ({TMF}), saudi arabia.
\newblock \emph{Processes}, 11\penalty0 (5):\penalty0 1545, May 2023.
\newblock \doi{10.3390/pr11051545}.
\newblock URL \url{https://doi.org/10.3390/pr11051545}.

\bibitem[Ifrene et~al.(2023)Ifrene, Egenhoff, Pothana, Nagel, and
  Li]{Ifrene2023}
Ghoulem Ifrene, Sven Egenhoff, Prasad Pothana, Neal Nagel, and Bo~Li.
\newblock High-resolution fluid flow simulation in x-crossing rough fractures.
\newblock 06 2023.
\newblock \doi{10.23265/ARMA-2023-0333}.

\bibitem[Liu et~al.(2021)Liu, Zhang, Liang, Temizel, Basri, and
  Mesdour]{Liu2021}
Hui~Hai Liu, Jilin Zhang, Feng Liang, Cenk Temizel, Mustafa~A. Basri, and Rabah
  Mesdour.
\newblock Incorporation of physics into machine learning for production
  prediction from unconventional reservoirs: A brief review of the gray-box
  approach, 11 2021.
\newblock ISSN 10946470.

\bibitem[Bikmukhametov and J\"{a}schke(2020)]{Bikmukhametov2020}
Timur Bikmukhametov and Johannes J\"{a}schke.
\newblock First principles and machine learning virtual flow metering: A
  literature review.
\newblock \emph{Journal of Petroleum Science and Engineering}, 184:\penalty0
  106487, January 2020.
\newblock \doi{10.1016/j.petrol.2019.106487}.
\newblock URL \url{https://doi.org/10.1016/j.petrol.2019.106487}.

\bibitem[Zhang et~al.(2022{\natexlab{c}})Zhang, He, AlSinan, Li, Kwak, and
  Hoteit]{ZhenZhang2022b}
Zhen Zhang, Xupeng He, Marwah AlSinan, Yiteng Li, Hyung Kwak, and Hussein
  Hoteit.
\newblock Deep learning model for co2 leakage detection using pressure
  measurements.
\newblock volume 2022-October. Society of Petroleum Engineers (SPE),
  2022{\natexlab{c}}.
\newblock ISBN 9781613998595.
\newblock \doi{10.2118/209959-MS}.

\bibitem[Aminu et~al.(2017)Aminu, Nabavi, Rochelle, and Manovic]{Aminu2017}
Mohammed~D. Aminu, Seyed~Ali Nabavi, Christopher~A. Rochelle, and Vasilije
  Manovic.
\newblock A review of developments in carbon dioxide storage.
\newblock \emph{Applied Energy}, 208:\penalty0 1389--1419, December 2017.
\newblock \doi{10.1016/j.apenergy.2017.09.015}.
\newblock URL \url{https://doi.org/10.1016/j.apenergy.2017.09.015}.

\bibitem[Krishnapriyan et~al.(2021)Krishnapriyan, Gholami, Zhe, Kirby, and
  Mahoney]{Krishnapriyan2021}
Aditi~S. Krishnapriyan, Amir Gholami, Shandian Zhe, Robert~M. Kirby, and
  Michael~W. Mahoney.
\newblock Characterizing possible failure modes in physics-informed neural
  networks, 2021.

\bibitem[Wang et~al.(2020)Wang, Teng, and Perdikaris]{Wang2020b}
Sifan Wang, Yujun Teng, and Paris Perdikaris.
\newblock Understanding and mitigating gradient pathologies in physics-informed
  neural networks, 2020.

\bibitem[Robbins and Monro(1951)]{Robbins1951}
Herbert Robbins and Sutton Monro.
\newblock A stochastic approximation method.
\newblock \emph{The Annals of Mathematical Statistics}, 22\penalty0
  (3):\penalty0 400--407, September 1951.
\newblock \doi{10.1214/aoms/1177729586}.
\newblock URL \url{https://doi.org/10.1214/aoms/1177729586}.

\bibitem[Kingma and Ba(2017)]{Kingma2017}
Diederik~P. Kingma and Jimmy Ba.
\newblock Adam: A method for stochastic optimization, 2017.

\bibitem[Liu and Nocedal(1989)]{Liu1989}
Dong~C. Liu and Jorge Nocedal.
\newblock On the limited memory {BFGS} method for large scale optimization.
\newblock \emph{Mathematical Programming}, 45\penalty0 (1-3):\penalty0
  503--528, August 1989.
\newblock \doi{10.1007/bf01589116}.
\newblock URL \url{https://doi.org/10.1007/bf01589116}.

\bibitem[Schmidt et~al.(2021)Schmidt, Schneider, and Hennig]{Schmidt2021}
Robin~M. Schmidt, Frank Schneider, and Philipp Hennig.
\newblock Descending through a crowded valley - benchmarking deep learning
  optimizers, 2021.

\bibitem[Pascanu et~al.(2014)Pascanu, Dauphin, Ganguli, and
  Bengio]{Pascanu2014}
Razvan Pascanu, Yann~N. Dauphin, Surya Ganguli, and Yoshua Bengio.
\newblock On the saddle point problem for non-convex optimization, 2014.

\bibitem[Choromanska et~al.(2015)Choromanska, Henaff, Mathieu, Arous, and
  LeCun]{Choromanska2015}
Anna Choromanska, Mikael Henaff, Michael Mathieu, Gérard~Ben Arous, and Yann
  LeCun.
\newblock The loss surfaces of multilayer networks, 2015.

\bibitem[Lee et~al.(2017)Lee, Panageas, Piliouras, Simchowitz, Jordan, and
  Recht]{Lee2017}
Jason~D. Lee, Ioannis Panageas, Georgios Piliouras, Max Simchowitz, Michael~I.
  Jordan, and Benjamin Recht.
\newblock First-order methods almost always avoid saddle points, 2017.

\bibitem[Dauphin et~al.(2014)Dauphin, Pascanu, Gulcehre, Cho, Ganguli, and
  Bengio]{Dauphin2014}
Yann Dauphin, Razvan Pascanu, Caglar Gulcehre, Kyunghyun Cho, Surya Ganguli,
  and Yoshua Bengio.
\newblock Identifying and attacking the saddle point problem in
  high-dimensional non-convex optimization, 2014.

\bibitem[Markidis(2021)]{Markidis2021}
Stefano Markidis.
\newblock The old and the new: Can physics-informed deep-learning replace
  traditional linear solvers?, 2021.

\bibitem[Hendrycks and Gimpel(2023)]{Hendrycks2023}
Dan Hendrycks and Kevin Gimpel.
\newblock Gaussian error linear units (gelus), 2023.

\bibitem[Al-Safwan et~al.(2021)Al-Safwan, Song, and Waheed]{Alsafwan2021}
Ali Al-Safwan, Chao Song, and Umair~Bin Waheed.
\newblock Is it time to swish? comparing activation functions in solving the
  helmholtz equation using physics-informed neural networks, 2021.

\bibitem[Eger et~al.(2019)Eger, Youssef, and Gurevych]{Eger2019}
Steffen Eger, Paul Youssef, and Iryna Gurevych.
\newblock Is it time to swish? comparing deep learning activation functions
  across nlp tasks, 2019.

\bibitem[Jagtap et~al.(2020)Jagtap, Kawaguchi, and Karniadakis]{Jagtap2020}
Ameya~D. Jagtap, Kenji Kawaguchi, and George~Em Karniadakis.
\newblock Adaptive activation functions accelerate convergence in deep and
  physics-informed neural networks.
\newblock \emph{Journal of Computational Physics}, 404:\penalty0 109136, mar
  2020.
\newblock \doi{10.1016/j.jcp.2019.109136}.
\newblock URL \url{https://doi.org/10.1016%2Fj.jcp.2019.109136}.

\bibitem[He et~al.(2015)He, Zhang, Ren, and Sun]{He2015}
Kaiming He, Xiangyu Zhang, Shaoqing Ren, and Jian Sun.
\newblock Delving deep into rectifiers: Surpassing human-level performance on
  imagenet classification.
\newblock \emph{CoRR}, abs/1502.01852, 2015.
\newblock URL \url{http://arxiv.org/abs/1502.01852}.

\bibitem[Bengio et~al.(2009)Bengio, Louradour, Collobert, and
  Weston]{Bengio2009}
Yoshua Bengio, J{\'{e}}r{\^{o}}me Louradour, Ronan Collobert, and Jason Weston.
\newblock Curriculum learning.
\newblock In \emph{Proceedings of the 26th Annual International Conference on
  Machine Learning}. {ACM}, June 2009.
\newblock \doi{10.1145/1553374.1553380}.
\newblock URL \url{https://doi.org/10.1145/1553374.1553380}.

\end{thebibliography}

\end{document}